\documentclass{article} % For LaTeX2e
\usepackage{iclr2022_conference,times}

\usepackage{amsmath,amsfonts,bm}

\def\eqref#1{equation~\ref{#1}}
\def\1{\bm{1}}

\DeclareMathAlphabet{\mathsfit}{\encodingdefault}{\sfdefault}{m}{sl}
\SetMathAlphabet{\mathsfit}{bold}{\encodingdefault}{\sfdefault}{bx}{n}

\usepackage[utf8]{inputenc} % allow utf-8 input
\usepackage[T1]{fontenc}    % use 8-bit T1 fonts
\usepackage{hyperref}       % hyperlinks
\usepackage{url}            % simple URL typesetting
\usepackage{booktabs}       % professional-quality tables
\usepackage{amsfonts}       % blackboard math symbols
\usepackage{nicefrac}       % compact symbols for 1/2, etc.
\usepackage{microtype}      % microtypography
\usepackage{xcolor}         % colors
\usepackage{smile}
\usepackage{tcolorbox}
\usepackage{wrapfig}
\newcommand{\ours}{{SAGE}}

\title{No Parameters Left Behind: Sensitivity Guided Adaptive Learning Rate for Training Large Transformer Models}

\author{Chen Liang*, Haoming Jiang$^\dagger$, Simiao Zuo*, Pengcheng He$^\star$, Xiaodong Liu$^\diamond$, \\ \textbf{Jianfeng Gao}$^\diamond$, \textbf{Weizhu Chen}$^\star$ \& \textbf{Tuo Zhao}* \\
* Georgia Institute of Technology, $^\dagger$ Amazon, $^\star$ Microsoft Azure AI, $^\diamond$ Microsoft Research\\
\texttt{\{cliang73,simiaozuo,tourzhao\}@gatech.edu} \\
\texttt{jhaoming@amazon.com} \\
\texttt{\{penhe,xiaodl,jfgao,wzchen\}@microsoft.com}
}

\iclrfinalcopy % Uncomment for camera-ready version, but NOT for submission.

\begin{document}
\maketitle
% !TEX encoding = UTF-8
% !TEX Root = main.tex
\begin{abstract} 
\label{sec:abstract}

Recent research has shown the existence of significant redundancy in large Transformer models. One can prune the redundant parameters without significantly sacrificing the generalization performance. However, we question whether the redundant parameters could have contributed more if they were properly trained. To answer this question, we propose a novel training strategy that encourages all parameters to be trained sufficiently. Specifically, we adaptively adjust the learning rate for each parameter according to its sensitivity, a robust gradient-based measure reflecting this parameter's contribution to the model performance. A parameter with low sensitivity is redundant, and we improve its fitting by increasing its learning rate. In contrast, a parameter with high sensitivity is well-trained, and we regularize it by decreasing its learning rate to prevent further overfitting. We conduct extensive experiments on natural language understanding, neural machine translation, and image classification to demonstrate the effectiveness of the proposed schedule. Analysis shows that the proposed schedule indeed reduces the redundancy and improves generalization performance.

\end{abstract}
%!TEX root = main.tex

% \vspace{-0.2in}
\section{Introduction}
\label{sec:intro}

% \vspace{-0.2in}

\begin{wrapfigure}{r}{0.3\textwidth}
    % \vspace{-0.15in}
    \centering
    \includegraphics[width=1\linewidth,height=0.8\linewidth]{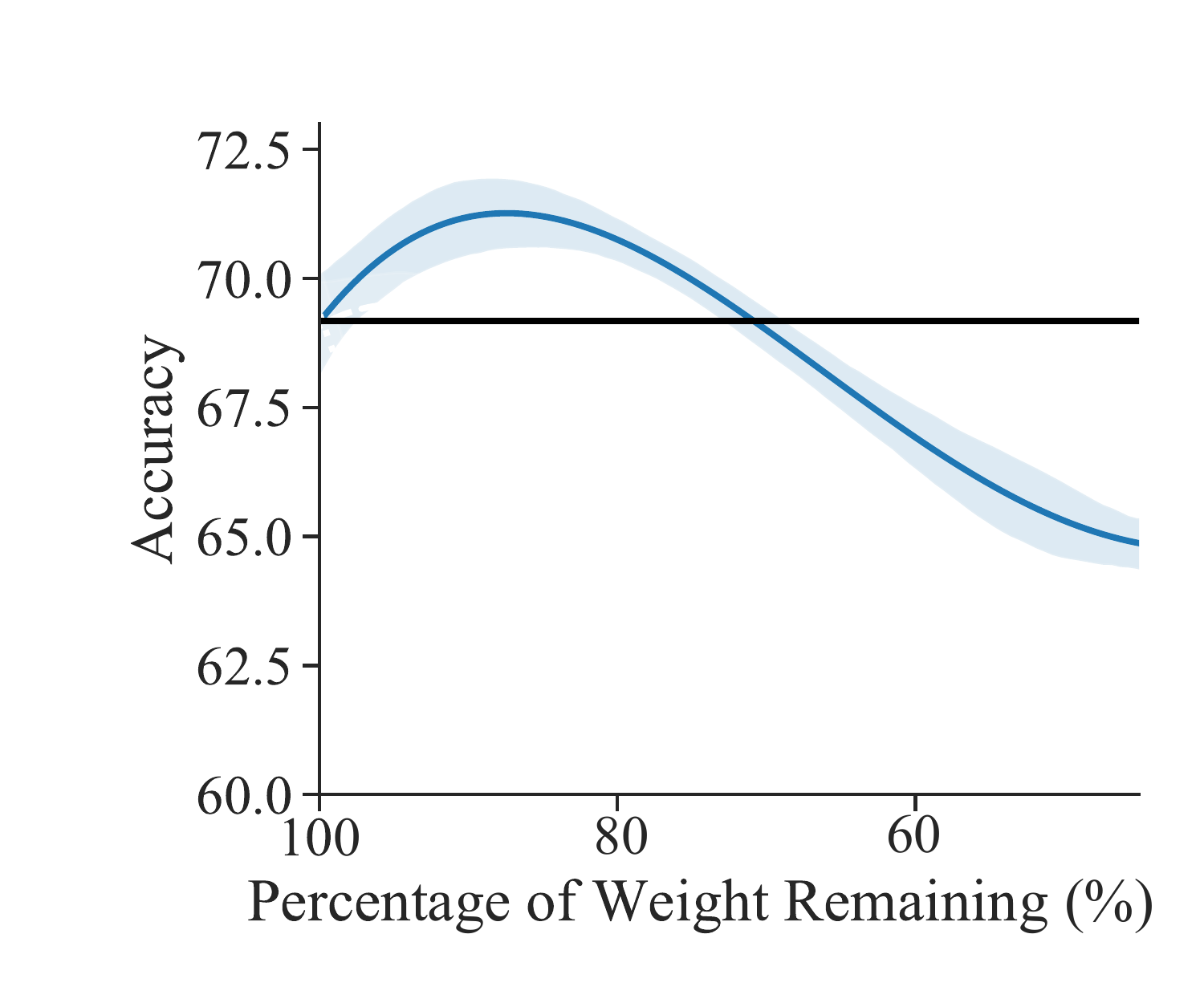}
	\caption{Validation results of fine-tuning BERT-base at different sparsity levels on the RTE dataset \citep{wang2018glue} in \citet{liang2021super}. Solid black curve represents the full model performance.}
	\label{fig:gen_ilst}
% 	\vspace{-0.15in}
\end{wrapfigure}

Large-scale Transformer models have achieved remarkable success in various fields. Performance of these models scales with their number of parameters, which can be up to hundreds of millions, e.g.,  BERT \citep{devlin2018bert}, DeBERTa \citep{he2020deberta}, GPT-3 \citep{brown2020language}. Recent research, however, has shown the existence of significant redundancy in the Transformer models \citep{michel2019sixteen, fan2019reducing, wang2019structured, chen2020lottery, sanh2020movement}. For example, \citet{sanh2020movement} removes around 90\% of the parameters, and the models exhibit only a marginal performance drop.

The existence of redundancy can hurt the model performance. Recent works have demonstrated that the removal of the redundant parameters can lead to better generalization performance, a phenomenon observed in both small-scale models \citep{mozer1989skeletonization, rasmussen2001occam, grunwald2007minimum} and large-scale Transformer models \citep{bartoldson2019generalization, voita2019analyzing, hou2020dynabert,liang2021super}. As illustrated in Figure~\ref{fig:gen_ilst}, with up to $20\%$ of the parameters pruned, the generalization performance boosts up to $1\%$.

% \citet{mozer1989skeletonization, rasmussen2001occam, grunwald2007minimum} have shown that redundancy elimination leads to better generalization in small models with tens to hundreds of parameters. In large-scale transformer models, %empirical results have also verified the existence of this phenomenon \citep{bartoldson2019generalization, voita2019analyzing, hou2020dynabert,liang2021super}.
% empirical results have also verified that removing the redundancy leads to better generalization performance, as illustrated in Figure~\ref{red_ilst} \citep{bartoldson2019generalization, voita2019analyzing, hou2020dynabert,liang2021super}.

As a result, we aim to improve model generalization through redundancy elimination. However, the existence of redundancy has long been regarded as inevitable. The common belief is that, in each network, there always exists a set of parameters ``born'' to be useless \citep{frankle2018lottery, liu2018rethinking}. Following this belief, pruning, where redundant parameters are directly zeroed out, becomes one of the most widely adopted solutions to redundancy elimination. However, we ask a critical question here:

%The existence of redundancy has long been regarded as inevitable. The common belief is that, in each network, there always exists a set of parameters ``born'' to be useless. For example, \citet{frankle2018lottery} suggest that poorly initialized parameters may not be well-optimized; \citet{liu2018rethinking} suggest that a suboptimal model architecture may result in specific structures being redundant. Following this belief, pruning, where redundant parameters are directly zeroed out, becomes one of the most widely adopted solutions to redundancy elimination. However, we ask a critical question here:

% \tz{Use your paper on super tickets as the major motivation}

\begin{tcolorbox}
\centering
\textit{Are these parameters really redundant, or just insufficiently trained by commonly used training strategies?}
\end{tcolorbox} 

Our question is motivated by empirical observations, which show that training strategies indeed play a role in causing redundancy. For example, different learning rates (Table~\ref{tb:mot}), random seeds and optimizers \citep{morcos2019one} can produce models with similar performance but different sets of redundant parameters. This suggests that the redundancy of parameters depends on the training strategy: 
%This suggests that redundant parameters can potentially become important once they are properly optimized.
A training strategy often prefers specific parameters and provides them with sufficient training. In contrast, the other parameters receive insufficient training and become under-fitted. As a result, these parameters become redundant, such that they fail to contribute to the generalization and prevent the model from achieving its ideal performance. Therefore, we hypothesize that with a desirable training strategy, these redundant parameters can receive more sufficient training and become useful ultimately. 

\begin{wraptable}{r}{0.35\textwidth}
%\vspace{-0.15in}
\begin{tabular}{c|c}
\toprule
\textbf{OVLP Among} & Avg \% OVLP \\ \midrule
2 Models &     59.8\%\\ 
3 Models &     46.5\%\\ 
5 Models &     35.7\%\\ \bottomrule
\end{tabular}
\caption{Percentage of overlapping between the $30\%$ most redundant parameters in $5$ BERT-base models fine-tuned using $\{1,5,8,10,20\}\times10^{-5}$ as learning rates on SST-2.}
\label{tb:mot}
%\vspace{-0.05in}
\end{wraptable}

We verify the hypothesis by proposing a novel training strategy, which encourages all parameters to be trained sufficiently. Throughout the training process, we simultaneously excite the under-fitted parameters to reduce redundancy and regularize the well-fitted parameters to prevent overfitting.

More specifically, we propose an adaptive learning rate schedule -- {\ours} (Sensitivity-guided Adaptive learninG ratE), where each parameter learns at its own pace guided by its sensitivity. Sensitivity originated in model pruning, where it is used to measure the redundancy of the parameters \citep{molchanov2016pruning, molchanov2019importance, theis2018faster,lee2018snip, ding2019global}. In pruning literature, parameters with low sensitivity are considered redundant. Since a redundant parameter could be insufficiently trained and under-fitted, we promote its training by increasing its learning rate. In contrast, for a parameter with high sensitivity, i.e., it is considered sufficiently trained and well-fitted, we slow down its training by decreasing its learning rate to prevent overfitting.

% %%%% Numerator intuition %%%%%%%%%%
% [stochastic, variation, uncertainty] + [proportional to ratio]
% To view sensitivity as a robust indicator of parameters' importance, we encourage its deviation through time to be small. As a result, we consider the local stability of a neuron's sensitivity as another guiding factor. Significant oscillation in a neuron's sensitivity implies it is fastly bouncing between being ``important'' or being ``redundant''. The neuron is possibly trapping in an unstable landscape. Therefore, we increase its learning rate to help it jump out and explore a flatter minimum. In contrast, a steady sensitivity implies the neuron is assured about its importance in the current model. Therefore, we update it carefully with a lower learning rate. Worth noticing, although a neuron may stabilize in the current model, it can be revitalized in the future, as other parameters will keep altering the landscape. 

Moreover, we introduce a local temporal variation of the sensitivity as a second factor to further guide the learning rate. The local temporal variation essentially measures the uncertainty of sensitivity, which mainly comes from two sources: (1) The sensitivity can have large variance due to data sampling.
This is because during training, the sensitivity is evaluated using a randomly sampled mini-batch instead of all the training data.
% we need to intensively measure sensitivity of all parameters.
%In each iteration, this is achieved alongside with the randomly sampled mini-batch, i.e., 
% To ease the scalability of each iteration, the sensitivity is evaluated using a randomly sampled mini-batch instead of all the training data.
% To ease the scalability of each iteration, we measure the sensitivity using a stochastic approximation involving only a randomly sampled 
% mini-batch of the training data, which potentially leads to a large variance;
(2) The sensitivity of a parameter may not be stable and can vary drastically among iterations, which introduces extra uncertainty. We define the local temporal variation of a parameter as the absolute difference between its sensitivity and an exponential moving average of its sensitivity from all previous iterations. A large local temporal variation implies high uncertainty in the sensitivity at the current iteration, and therefore it is not yet a reliable indicator of redundancy. Accordingly, we should avoid significantly decreasing its learning rate even though its sensitivity at the current iteration might be large.

Therefore, we eventually require the overall learning rate schedule for each parameter to be proportional to the ratio between the local temporal variation and the sensitivity. This can effectively account for the uncertainty issue in sensitivity.

% %%%%%% Experiments %%%%%%%%%%%
We conduct experiments on a wide range of tasks and models to demonstrate the effectiveness of {\ours}. In natural language understanding, the fine-tuning performance of BERT-base \citep{devlin2018bert} and RoBERTa-large \citep{liu2019roberta} improves $1.4$ and $0.6$ task-average score on the dev set of the GLUE benchmark \citep{wang2018glue}, respectively. Furthermore, {\ours} improves neural machine translation performance using Transformer-base \citep{vaswani2017attention} on two datasets, suggesting it also benefits training-from-scratch. %Our proposed method is generic to other types of data. %It improves the image classification accuracy by $0.7\%$ and $0.9\%$ on CIFAR100 \citep{krizhevsky2009learning} using ResNet-20 and PreResNet-110 \citep{he2016deep}, respectively.
{\ours} also boost the image classification accuracy on ImageNet dataset \citep{deng2009imagenet} with Vision Transformer models \citep{dosovitskiy2020image}. Furthermore, our experiments demonstrate {\ours} is complementary to various types of optimizers, e.g., SGD \citep{robbins1951stochastic}, Adam,  and Adamax \citep{kingma2014adam}.

% %%%%%% Analysis %%%%%%%%%%%
Moreover, we observe several favorable proprieties of {\ours}. First, it leads to balanced and sufficient training on all parameters and produces a better-generalized model. Second, {\ours} is complementary to state-of-the-art training methods. Specifically, we show that {\ours} achieves better performance on GLUE when combined with adversarial regularization~\citep{jiang2019smart}. Our code has been released at \url{https://github.com/cliang1453/SAGE}.
% Additionally, we carry out extensive analysis to verify that {\ours} leads to more balanced and more sufficient training on all parameters. Our proposed method is also complementary to state-of-the-art training methods. Specifically, we show that {\ours} achieves better performance on GLUE when combined with adversarial regularization~\citep{jiang2019smart}.

%!TEX root = main.tex
% \vspace{-0.05in}
\section{Preliminary}
\label{sec:background}
% \vspace{-0.05in}

We briefly review the sensitivity of the parameters and adaptive learning rate methods.

\subsection{Sensitivity of the parameters}

The sensitivity of a parameter essentially approximates the change in the loss magnitude when this parameter is completely zeroed-out \citep{lecun1990optimal,mozer1989skeletonization}. If the removal of a parameter causes a large influence on the loss, then the model is sensitive to it. More specifically, we define a deep neural network with parameters $\boldsymbol{\Theta} = [\theta_1, ...,\theta_J] \in \RR^{J}$, where for $j=1,...,J$, $\theta_j \in \RR$ denotes each parameter. We further define $\boldsymbol{\Theta}_{j, -j} = [0, ..., 0, \theta_j, 0, ..., 0] \in \RR^{J}$. We denote the loss of the model as $L(\boldsymbol{\Theta})$, and the gradients of the loss with respect to $\boldsymbol{\Theta}$ as $\nabla_{\boldsymbol{\Theta}} L(\boldsymbol{\Theta})$. The sensitivity of the $j$-th parameter is defined as the magnitude of the gradient-weight product:
\begin{align}
    I_j = |\boldsymbol{\Theta}_{j, -j}^{\top} \nabla_{\boldsymbol{\Theta}} L(\boldsymbol{\Theta})|.
    \label{eq:def}
\end{align} 
This definition is derived from the first-order Taylor expansion of $L(\cdot)$ with respect to $\theta_j$ at $\boldsymbol{\Theta}$. Specifically,  $I_j$ approximates the absolute change of the loss given the removal of $\theta_j$: 
\begin{align*}
    \boldsymbol{\Theta}_{j, -j}^{\top} \nabla_{\boldsymbol{\Theta}} L(\boldsymbol{\Theta})\approx  L(\boldsymbol{\Theta}) - L(\boldsymbol{\Theta} - \boldsymbol{\Theta}_{j, -j}).
\end{align*}
%where we assume the higher-order terms are negligible. 

The sensitivity was originally introduced for model pruning \citep{molchanov2016pruning,molchanov2019importance,theis2018faster,lee2018snip,ding2019global,xiao2019autoprune}, and it was commonly used as an ``importance score'' for model weights. The parameters with high sensitivity are of high importance and should be kept \citep{lubana2020gradient}. Parameters with low sensitivity are considered redundant, and they can be safely pruned with only marginal influence on the model loss.

% \vspace{-0.03in}
\subsection{Adaptive Learning Rate Methods}
\label{sec:adaptive_learning_rate}
% \vspace{-0.03in}

Adaptive learning rate methods adjust the learning rate of each individual parameter based on the training progress. %, and play an important role in training Transformer models. 
Most of these methods focus on adapting the training to the optimization landscape, e.g., AdaGrad \citep{duchi2011adaptive}, AdaDelta \citep{zeiler2012adadelta}, RMSProp \citep{hinton2012neural}, Adam\citep{kingma2014adam} and RAdam \citep{liu2019variance}. Their purpose is to make the model converge faster to the first-order stationary solutions. Specifically, these methods prefer updating the weights with smaller second-order moments, as the loss function is generally flat along directions corresponding to such weights. %However, the convergence guarantees of these methods are usually established upon Lipschitz continuous gradient and bounded stochastic gradient assumptions, which actually do not hold for Transformer models in practice. Moreover, such convergence to the first-order stationary solutions does not imply any theoretical guarantees on the model generalization performance. 

%While improving generalization performance becomes a critical challenge for Transformer models, there are few adaptive learning rate methods proposed for this purpose. 
There are also some adaptive learning rate methods focusing on the perspective of improving model generalization \citep{loshchilov2018fixing,foret2020sharpness}. For example, AdamW \citep{loshchilov2018fixing} propose to decouple the weight decay and gradient update to avoid regularizing weights that have larger gradient magnitudes with a weaker strength.

\section{Method}
\label{sec:method}
%\vspace{-0.05in}

We introduce our proposed adaptive learning rate schedule, {\ours}. Our method customizes a specific learning rate for each parameter at each iteration. A parameter's learning rate at a certain iteration is determined by two factors: sensitivity and its local temporal variation.

\textbf{Sensitivity of the parameters.} At the $t$-th iteration, following Eq.~(\ref{eq:def}), we define the sensitivity of $\theta_j^{(t)}$ as
\begin{align}
	I_j^{(t)} = |\boldsymbol{\Theta}_{j, -j}^{(t)\top} \nabla_{\boldsymbol{\Theta}^{(t)}} L(\boldsymbol{\Theta}^{(t)})|,
\label{eq:ipt}
\end{align}
which reflects the influence of removing $\theta_j^{(t)}$ in the model loss. In previous literature, $\theta_j^{(t)}$ is considered redundant when $I_j^{(t)}$ is small. In contrast, we hypothesize that $\theta_j^{(t)}$ is just insufficiently trained and under-fitted, and can become less redundant when receiving further training.

\textbf{Local temporal variation.} Recall that the sensitivity measure involves excessive uncertainty, which comes from: (1) Sensitivity is measured based on a randomly sampled mini-batch of the training data at each iteration, which leads to a large variance; (2) Sensitivity can be unstable and vary drastically, as changes of the model introduce extra uncertainty to the measure. 

One way to measure the uncertainty of sensitivity of $\theta_j$ is the absolute change of sensitivity, i.e., $|I_j^{(t)} - I_j^{(t-1)}|$. Such a quantity often has a large variance in practice. Therefore, we propose to keep track of an exponential moving average of $I_j^{(t)}$ as
\begin{align*}
	\hat{I}_j^{(t)} = \beta_{0} \hat{I}_j^{(t-1)} + (1-\beta_{0}) I_j^{(t)},
\end{align*} 
where $\hat{I}_j^{(0)} = 0$ and $\beta_{0}\in (0,1)$ is a hyper-parameter. Based on $\hat{I}_j^{(t)}$, we measure the uncertainty of the $j$-th parameter's sensitivity using the local temporal variation defined as:
\begin{align}
	U_j^{(t)} = |I_j^{(t)} - \hat{I}_j^{(t)}|.
	\label{eq:std}
\end{align}
We remark that a large $U_j^{(t)}$ implies that there exists high uncertainty in $I_j^{(t)}$, and therefore it is not yet a reliable indicator of the redundancy of $\theta_j^{(t)}$.

\textbf{Algorithm.} We denote the learning rate at the $t$-th iteration as $\eta^{(t)}$ under the original schedule. Then the sensitivity-guided learning rate for the $j$-th parameter at the $t$-th iteration can be computed as
\begin{align}
	\eta^{(t)}_j = \eta^{(t)} \cdot \frac{U_j^{(t)} + \epsilon}{\hat{I}_j^{(t)} + \epsilon} = \eta^{(t)} \cdot \frac{|I_j^{(t)} - \hat{I}_j^{(t)}| + \epsilon}{\hat{I}_j^{(t)} + \epsilon}, 
	\label{eq:alg}
\end{align} 
where $0 < \epsilon \ll 1$ prevents zero learning rate and zero denominator. Algorithm~\ref{alg:main} shows the {\ours} algorithm for SGD, and extensions to other algorithms, such as Adam~\citep{kingma2014adam}, are straightforward (Appendix~\ref{app:sage-adam}).

In Eq.~(\ref{eq:alg}), we place $\hat{I}_j^{(t)}$ in the denominator, as one of our goals is to encourage all parameters to be sufficiently trained. If $\hat{I}_j^{(t)}$ is small, we promote its training by increasing its learning rate. If $\hat{I}_j^{(t)}$ is large, we slow down its training to prevent overfitting by decreasing its learning rate.

We place $U_j^{(t)}$ in the numerator to measure the uncertainty in the sensitivity. A large $U_j^{(t)}$ implies $I_j^{(t)}$ is not yet a reliable indicator of the redundancy in $\theta_j^{(t)}$. We thus avoid significantly decreasing its learning rate.

% Sincethe chance of getting consecutive small or large learning rate is small.

%Since the learning rate is modulated at individual parameters, its variance can be large. Ideally, the algorithm allows all parameters to quickly escape from extreme learning rates. For example, a parameter with stable and high sensitivity may have a close to zero learning rate, and thus its weight value is likely stuck at the current step. However, as the rest of the parameters keep changing the global landscape, the learning rate will soon change along with its sensitivity. In practice, to prevent training instability, we still take precautions for such extreme cases. Specifically, we adjust \eqref{eq:alg1} as:
% \begin{align}
% 	\eta^{(t)}_j = \eta^{(t)} \cdot \frac{U_j^{(t)} + \epsilon}{\hat{I}_j^{(t)} + \epsilon}, 
% 	\label{eq:alg}
% \end{align} 

\begin{algorithm}[htb!]
	\caption{SGD-{\ours} ($\odot$ denotes Hadamard product and $\oslash$ denotes Hadamard division)}
	\label{alg:main}
	\begin{algorithmic}[1]
		\INPUT Model parameters $\boldsymbol{\Theta} \in \RR^{J}$; Data $\cD$; Learning rate schedule $\eta(\cdot)$; Total training iteration $T$; Moving average coefficient $\beta_{0}$. \\
		Initialize $\hat{I}^{(0)} = \boldsymbol{0} \in \RR^{J}$. 
		\For{$t = 1, ..., T$}
    		\State Sample a minibatch $b^{(t)}$ from $\cD$.
    		\State Compute gradient $\nabla_{\boldsymbol{\Theta^{(t)}}} L(b^{(t)}, \boldsymbol{\Theta}^{(t)})$. 
		\State $I^{(t)} = |\boldsymbol{\Theta}^{(t)} \odot \nabla_{\boldsymbol{\Theta^{(t)}}} L(b^{(t)}, \boldsymbol{\Theta}^{(t)})|$. 
		\State $\hat{I}^{(t)} = \beta_0 \hat{I}^{(t-1)} + (1-\beta_0) I^{(t)}$.
		\State $U^{(t)} = |I^{(t)} - \hat{I}^{(t)}|$.
    	\State $\boldsymbol{\Theta}^{(t+1)} = \boldsymbol{\Theta}^{(t)} - \eta^{(t)} (U^{(t)} + \epsilon) \oslash (\hat{I}^{(t)} + \epsilon) \odot \nabla_{\boldsymbol{\Theta^{(t)}}} L(b^{(t)}, \boldsymbol{\Theta}^{(t)})$.
		\EndFor
	\end{algorithmic}
\end{algorithm}

% \textbf{Compared with existing adaptive gradient-based algorithms.} %We further remark that our proposed schedule is complementary to existing adaptive methods, which \\

\textbf{Computation and memory usage.} {\ours} adds a marginal cost to computation and memory usage. At each iteration, we only perform an extra element-wise multiplication between the weight matrix and the corresponding gradient matrix obtained through back-propagation. The only memory cost is to store the exponential moving average of sensitivity.

%\textbf{Remark.} {\ours} aims to improve generalization through eliminating the weight redundancy. The quantities of our interest (i.e., Eq.~(\ref{eq:ipt}) and Eq.~(\ref{eq:std})) are related to the weight redundancy. They are not directly related to the moduli of the objective function, e.g., smoothness, curvature (which are of the interests for optimization). This is very different from the mainstream adaptive learning rate methods, which focus on the optimization aspect, i.e., faster convergence to the stationary solutions.

%\tz{Add a discussion here: The quantities of our interest in the learning rate schedule is about the weight redundancy. They are not directly related to the moduli of the  objective function, e.g., smoothness, curvature, which are of the interests for optimization.}
%%%%%%%%%%%%%%%%%%%%%%%%%%%%%%%%%%%%%
%\vspace{-0.05in}
\section{Experiments}
\label{sec:exp}
%\vspace{-0.05in}

We evaluate {\ours} on widely used benchmarks for natural language understanding (NLU), neural machine translation (NMT), and image classification.

%%%%%%%%%%%%%%%%%%%%%%%%%%%%%%%%%%%%%
%\vspace{-0.03in}
\subsection{Natural Language Understanding}
%\vspace{-0.03in}

\textbf{Model and data.} We evaluate the fine-tuning performance of the pre-trained language models, BERT-base \citep{devlin2018bert} and RoBERTa-large \citep{liu2019roberta}, on the General Language Understanding Evaluation (GLUE,~\citet{wang2018glue}) benchmark. GLUE contains nine NLU tasks, including textual entailment, question answering, sentiment analysis, and text similarity. Details about the benchmark are deferred to Appendix~\ref{app:glue-data}.

\textbf{Implementation Details.} We implement our method using the MT-DNN code-base\footnote{https://github.com/namisan/mt-dnn}. We follow the suggested training and hyper-parameters settings from \citet{liu2020mtmtdnn}. Specifically, we adopt Adam and Adamax \citep{kingma2014adam} with corrected weight decay \citep{loshchilov2018fixing} as the baseline optimizer and we set $\beta = (0.9, 0.999)$. We use a linear-decay learning rate schedule, and we apply {\ours} to both Adam and Adamax.

We select learning rates in range of $\{1,2,3,5,8\} \times \{10^{-5}, 10^{-4}\}$. We select $\beta_{0}$ in range of $[0.6, 0.9]$ with an increment of $0.05$. Other training details are reported in Appendix~\ref{app:glue-training}. 

\textbf{Main results.} Table~\ref{tb:glue-dev-results} and Table~\ref{tb:glue-test-results} show the evaluation results on the GLUE benchmark. The dev results are averaged over $5$ different random seeds, and all gains are statistically significant\footnote{The dev results on RoBERTa-large are averaged over $3$ different random seeds. All results have passed a paired student t-test with p-values less than $0.05$. The detailed statistics are summarized in Appendix~\ref{app:glue-exp}.}. We select the best single task model for test evaluation. 

\newcolumntype{C}{@{\hskip3pt}c@{\hskip3pt}}
\begin{table*}[htb!]
% \vspace{-0.05in}
\centering
\scriptsize
\begin{tabular}{l|l|CCCCCCCCCC}
\toprule
\multirow{2}{*}{\textbf{Model}} & \multirow{2}{*}{\textbf{Optimizer}} & \textbf{RTE} & \textbf{MRPC} & \textbf{CoLA} & \textbf{SST-2} & \textbf{STS-B} & \textbf{QNLI} & \textbf{QQP} & \textbf{MNLI-m/mm} & \textbf{Average} \\
&& Acc & Acc/F1 & Mcc & Acc & P/S Corr & Acc & Acc/F1 & Acc & Score \\\midrule
% \multicolumn{10}{l}{BERT\textsubscript{BASE}} \\\midrule
\multirow{5}{*}{BERT\textsubscript{BASE}} & \citet{devlin2018bert}   & - & -/86.7 & - & 92.7 & -/- & 88.4 & -/- & 84.4/- & - \\ \cmidrule(r){2-11}
& Adam   & 63.5 & 84.1/89.0 & 54.7 & 92.9 & 89.2/88.8 & 91.1 & 90.9/88.1 & 84.5/84.4 & 81.5 \\ 
& Adam-{\ours}   &\textbf{73.3} & \textbf{87.0/90.9} & \textbf{60.3} & \textbf{93.5} & \textbf{90.3/89.9} & \textbf{91.7} & \textbf{91.2/88.1} & \textbf{84.7/84.8} & \textbf{84.0} \\ \cmidrule(r){2-11}
& Adamax & 69.2 & 86.2/90.4 & 57.8 & 92.9 & 89.7/89.2 & 91.2 & 90.9/88.0 & 84.5/84.4 & 82.8 \\ 
& Adamax-{\ours} & \textbf{74.0} & \textbf{87.3/91.0} & \textbf{59.7} & \textbf{93.8} & \textbf{90.3/89.8} & \textbf{91.8} & \textbf{91.2/88.2} & \textbf{85.0/85.2} &  \textbf{84.2} \\ \midrule
%RoBERTa\textsubscript{LARGE} (Reported) & 86.6 & -/90.9 & 68.0 & 96.4 & 92.4/- & 94.7 & 92.2/- & 90.2/90.2 & \\ 
\multirow{3}{*}{RoBERTa\textsubscript{LARGE}} & \citet{liu2019roberta} & 86.6 & -/90.9 & 68.0 & 96.4 & 92.4/- & 94.7 & 92.2/- & 90.2/90.2 & - \\ \cmidrule(r){2-11}
& Adamax & 86.6 & 90.4/93.1 & 67.5 & 96.4 & 92.4/92.2 & 94.7 & 92.1/89.3 & 90.4/90.3 & 88.7 \\ 
& Adamax-{\ours} & \textbf{87.8} & \textbf{91.5/93.9} & \textbf{68.7} & \textbf{96.7} & \textbf{92.7/92.4} & \textbf{94.9} & \textbf{92.2/89.4} & \textbf{90.8/90.4} & \textbf{89.3} \\ \bottomrule
\end{tabular}
\caption{Single task fine-tuning dev results on GLUE. All results are from our implementations. `-' denotes missing results.}
\label{tb:glue-dev-results}
%\vspace{-0.05in}
\end{table*}

Our method gains $1.4$ on dev and $1.1$ on test of the task-average score on BERT-base. In large datasets, i.e., MNLI (392K) and QNLI (108K), {\ours} improves around $0.5$ points. In small datasets, i.e., RTE (2.5K) and CoLA (8.5K), we obtain more than $2$ points of improvements. Such observations indicate that {\ours} is very effective on the small datasets. Furthermore, {\ours} improves upon RoBERTa-large by $0.6$ average scores, suggesting {\ours} can still achieve significant improvements for larger and more adequately pre-trained models than BERT-base.
% Even with a more sufficiently pre-trained model, RoBERTa-large, {\ours} can still achieve significant improvement in the fine-tuning stage, by encouraging sufficient training .

\newcolumntype{C}{@{\hskip3pt}c@{\hskip3pt}}
\begin{table*}[htb!]
% \vspace{-0.05in}
\centering
\scriptsize
\begin{tabular}{@{\hskip6pt}l@{\hskip6pt}|CCCCCCCCCC}
\toprule
& \textbf{RTE} & \textbf{MRPC} & \textbf{CoLA} & \textbf{SST-2} & \textbf{STS-B} & \textbf{QNLI} & \textbf{QQP} & \textbf{MNLI-m/mm} & \textbf{Average} \\
& Acc & F1 & Mcc & Acc & P/S Corr & Acc & F1 & Acc & Score \\\midrule
BERT\textsubscript{BASE} \citep{devlin2018bert}  & 66.4 & 88.9   & 52.1 & 93.5 & 85.8  & 90.5 &  71.2   &  84.6/83.4 &  79.6 \\ 
BERT\textsubscript{BASE}, Adamax                 & 66.8 & 88.6   & 54.0 & 93.4 & 86.6  & 90.6 &  71.1   &  84.7/83.6 &  79.9 \\ 
BERT\textsubscript{BASE}, Adamax-{\ours}  & \textbf{69.8} & \textbf{89.7} & \textbf{54.5} & \textbf{94.1} & \textbf{87.1} & \textbf{90.8} & \textbf{71.3} & \textbf{84.9/83.8} & \textbf{80.7} \\ \bottomrule
\end{tabular}
\caption{Single task fine-tuning test results from the GLUE evaluation server.}
\label{tb:glue-test-results}
%\vspace{-0.05in}
\end{table*}

\newcolumntype{C}{@{\hskip4pt}c@{\hskip4pt}}
\begin{table*}[htb!]
%\vspace{-0.05in}
\centering
\small
\begin{tabular}{l|l|CC}
\toprule
\textbf{Model} & \textbf{Optimizer} & \textbf{IWSLT'14 De-En} & \textbf{WMT'16 En-De} \\ \midrule 
\multirow{2}{*}{Transformer\textsubscript{BASE}} & Adam & 34.5 & 27.3 \\ 
& Adam-{\ours} & \textbf{35.1} & \textbf{27.7} \\ \bottomrule
\end{tabular}
% \vspace{-0.05in}
\caption{Neural machine translation BLEU scores on test set. All results are from our implementation.}
\label{tb:nmt-results}
%\vspace{-0.05in}
\end{table*}

\newcolumntype{C}{@{\hskip4pt}c@{\hskip4pt}}
\begin{table*}[htb!]
%\vspace{-0.05in}
\centering
\small
\begin{tabular}{l|l|CC}
\toprule
\textbf{Model} & \textbf{Optimizer} & \textbf{CIFAR100} & \textbf{ImageNet}\\ \midrule 
\multirow{2}{*}{ViT-B/32} & SGD$^{*}$   &  91.97 &  81.28 \\ 
& SGD-{\ours}  & \textbf{92.68} & \textbf{81.72} \\ \midrule
\multirow{2}{*}{ViT-L/32} & SGD$^{*}$  & 93.04 & 80.99 \\ 
& SGD-{\ours} & \textbf{93.74} & \textbf{81.90} \\ \bottomrule
\end{tabular} 
%\vspace{-0.05in}
\caption{Image classification test accuracy. Results with $*$ are from \citet{dosovitskiy2020image}. ViT-B/32 and ViT-L/32 each denotes ViT-base and ViT-large model with $32\times 32$ input patch size.}
\label{tb:cls-test}
%\vspace{-0.15in}
\end{table*}

%%%%%%%%%%%%%%%%%%%%%%%%%%%%%%%%%%%%%%%%
%\vspace{-0.03in}
\subsection{Neural Machine Translation}
%\vspace{-0.03in}

\textbf{Model and Data.} We evaluate {\ours} on the Transformer-base NMT models \citep{vaswani2017attention} using two widely used NMT datasets, IWSLT'14 De-En \citep{cettolo2015iwslt}\footnote{https://wit3.fbk.eu/} and WMT'16 En-De \citep{bojar2016findings}\footnote{http://data.statmt.org/wmt16/translation-task/}. IWSLT'14 De-En is a low-resource dataset, which contains 160K sentence pairs. WMT'16 En-De is a rich-resource dataset, which contains 4.5M sentence pairs. Dataset and pre-processing details are deferred to Appendix~\ref{app:nmt-data}.

\textbf{Implementation Details.} We implement the algorithms using the \textit{fairseq} code-base and follow the training and hyper-parameters settings from \citet{ott2018scaling, ott2019fairseq}. Specifically, we adopt the inverse square root learning rate schedule and we employ Adam \citep{kingma2014adam} as the optimizer with $\beta=(0.9, 0.98)$. We apply {\ours} to the same setting.  

We select learning rates in range of $\{5, 7\}\times 10^{-5} \cup \{1, 2\}\times 10^{-4}$ and select $\beta_{0}$ in range of $\{0.5, 0.6, 0.7, 0.8, 0.9\}$. Comprehensive training details are reported in Appendix~\ref{app:nmt-training}.

\textbf{Main results.} Table~\ref{tb:nmt-results} shows the BLEU scores on the IWSLT'14 De-En and the WMT'16 En-De test set, where {\ours} improves around $0.6$ and $0.4$ points, respectively. This suggests that other than fine-tuning, {\ours} can also improve the generalization of trained-from-scratch models in both low-resource and rich-resource settings. 
%For WMT'16 En-De, we report the sacreBLEU \citep{post2018call} with compound splitting \footnote{The tokenizer version is tok.13a+version.1.5.1.}. 

%%%%%%%%%%%%%%%%%%%%%%%%%%%%%%%%%%%%%%%%
%\vspace{-0.03in}
\subsection{Image Classification}

\textbf{Model and data.} We evaluate {\ours} using Vision Transformer models (ViT) on the CIFAR100 \citep{krizhevsky2009learning} and ILSVRC-2012 ImageNet dataset \citep{deng2009imagenet}. Specifically, we evaluate the fine-tuning performance of the ViT-base and ViT-large pre-trained using ImageNet-21k, a superset of ImageNet dataset with $21$k classes and $14$M images. Data and pre-processing details are deferred to Appendix~\ref{app:img-data}. %We also evaluate {\ours} on convolutional networks, e.g., ResNet-20 (9 Basic Blocks, \citet{he2016deep}), PreResNet-110 (54 Basic Blocks, \citet{he2016identity}), and XSE-ResNeXt-50 \citep{he2019bag}, and the results are reported in Appendix~\ref{app:img-add-exp}.

\textbf{Implementation details.} All experiments follow the suggested training configuration of \citet{dosovitskiy2020image} and a jax-implemented code base \footnote{https://github.com/google-research/vision\_transformer}. We adopt SGD as the baseline optimizer with a momentum factor $0.9$. We fine-tune the models for $100$K steps for CIFAR100, and $200$K steps for ImageNet. We select learning rates in range of $\{0.02, 0.05, 0.08, 0.1\}$ and select $\beta_0$ in range of $\{0.85, 0.90, 0.95\}$. Comprehensive training details are reported in Appendix~\ref{app:img-training}.

\textbf{Main results.}
Table~\ref{tb:cls-test} shows the evaluation results on CIFAR100 and ImageNet. {\ours} outperforms baselines by a significant margin. This demonstrates that {\ours} is quite general, and can be applied to various tasks (e.g., NLP and computer vision) and optimizers (e.g., Adam, Adamax and SGD).

%\vspace{-0.05in}
\section{Analysis}
\label{sec:anls}
%\vspace{-0.05in}
We verify that {\ours} leads to more sufficient training (Section~\ref{sec:suff-train}), better generalization performance (Section~\ref{sec:ana_generalization}), and is complementary to existing state-of-the-art regularization methods (Section~\ref{sec:sota_combine}). We also provide ablation studies in Appendix~\ref{app:ana_ablation}.

%\vspace{-0.03in}
\subsection{{\ours} Leads to More Sufficient Training}
\label{sec:suff-train}
%\vspace{-0.03in}

Recall that {\ours} adjusts the learning rate for each parameter according to two factors: the sensitivity of parameters and the local temporal variation of sensitivity. By inspecting these factors, we verify that {\ours} leads to more sufficient training.

\begin{figure}[htb!]
    %\vspace{-0.05in}
    \centering
    \includegraphics[width=1.0\linewidth]{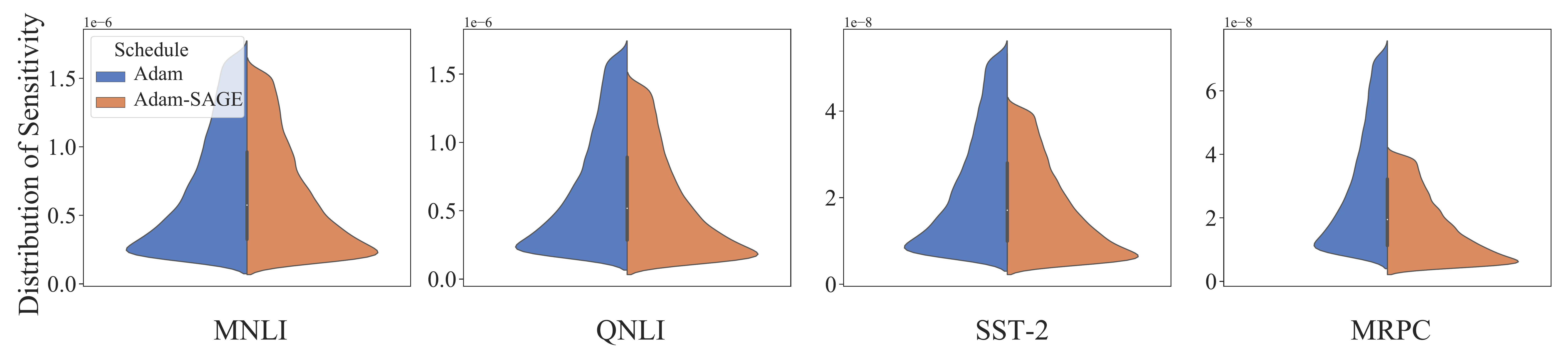}
	%\vspace{-0.1in}
	\caption{The sensitivity distribution of the BERT-base models fine-tuned on GLUE tasks. Note that we drop some outliers to ease visualization.}
	\label{fig:ipt_dist}
	%\vspace{-0.05in}
\end{figure}

\textbf{The sensitivity distribution is more concentrated.} Figure~\ref{fig:ipt_dist} shows the sensitivity distribution of parameters in the {\ours} optimized models and the baseline models. We select the hyper-parameters that yield the best generalization performance on the BERT-base model, and we evaluate the sensitivity of each parameter using the entire training set. See Appendix~\ref{app:suff-train} for implementation details.

We observe that the sensitivity distribution exhibits a lower variance in the {\ours} optimized models than the baseline models. This suggests that the sensitivity of parameters becomes more concentrated. In other words, the amount of each parameter's contribution is more balanced, and the model is more sufficiently trained. %Sensitivity is a metric for **comparing** parameters within a single model. The absolute magnitude does not imply anything about the redundancy, or the accuracy of sensitivity estimates. Comparing the magnitude of weights sensitivity across models is not very meaningful, as different models’ weights can be of different scales. However, the distribution of weights sensitivity is more condense, which agrees with our intention.

%$$\bullet~$ The sensitivity distribution exhibits a lower mean in the {\ours} optimized model. The baseline model contains a larger number of highly sensitive parameters than the {\ours} optimized model. This suggests the baseline model maybe over-fitted, as a small perturbation on highly sensitive parameters can drastically influence the model performance. In contrast, {\ours} regularizes highly sensitivity parameters and alleviate overfitting.

%$\bullet~$ The variance and the mean of the sensitivity distribution decreases along with the task size. For example, {\ours} leads to a more concentrated and downward-shifted distribution in MRPC (3.5K) than in MNLI (392K). This is because training on smaller tasks leads to more severe over-fitting, and {\ours} resolves overfitting through regularizing unbalanced sensitivity.

\begin{figure}[htb!]
    %\vspace{-0.15in}
    \centering
    \includegraphics[width=0.9\linewidth]{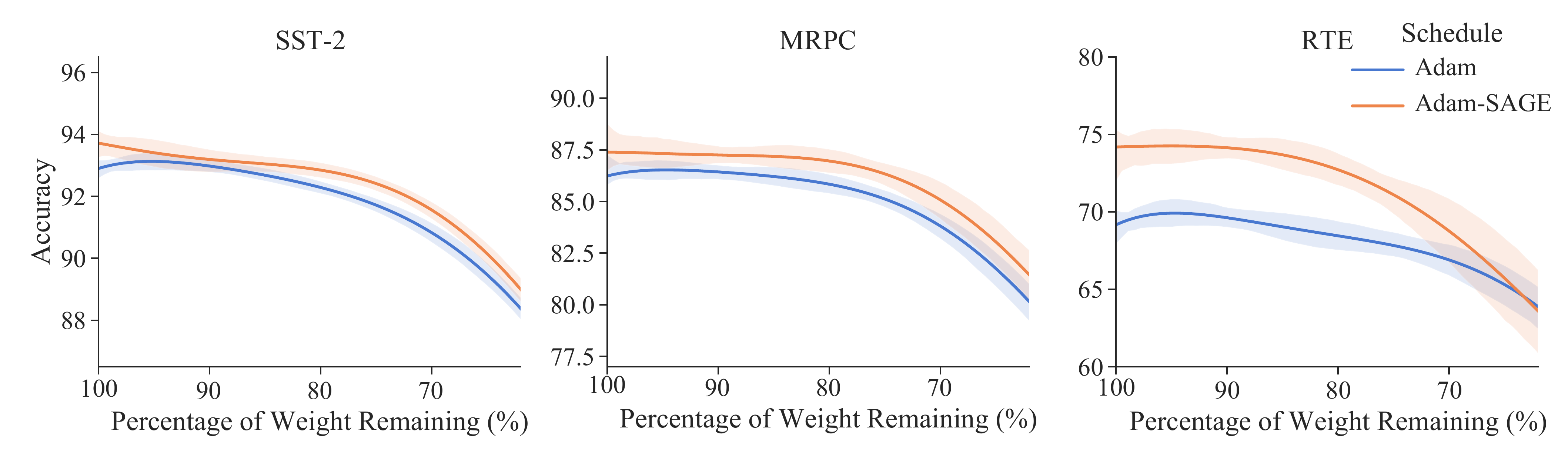}
    \includegraphics[width=0.9\linewidth]{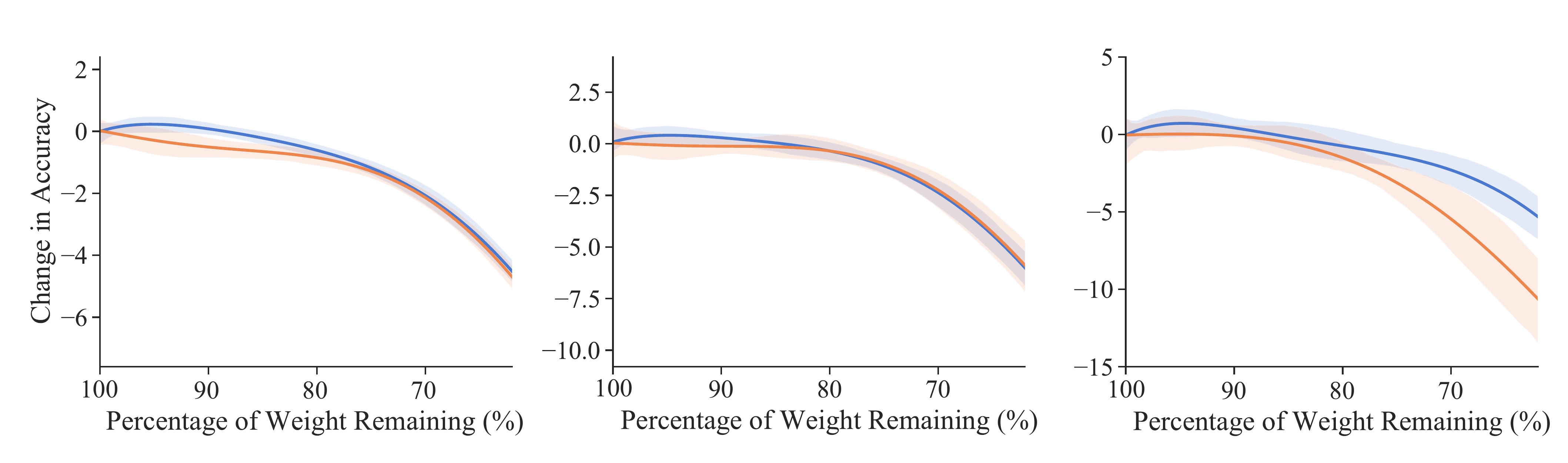}
	%\vspace{-0.1in}
	\caption{\textit{Upper}: Model generalization performance at different pruning ratios; \textit{Lower}: Change in generalization performance with respect to the full model. Pruning is conducted on the fine-tuned BERT-base models.}
	\label{fig:ipt_pruning}
	%\vspace{-0.15in}
\end{figure}

\textbf{Even the most redundant parameters contribute to the model performance.} Recall that sensitivity is a type of importance score in pruning, which is a straightforward approach to measure each parameter's contribution. Therefore, we conduct an unstructured, one-shot pruning experiment on the fine-tuned BERT-base models. Specifically, we remove up to $40\%$ parameters\footnote{Embedding weights are excluded.} with the lowest sensitivity scores and evaluate the pruned models' performance. We average the results over $5$ models trained with different random seeds. Figure~\ref{fig:ipt_pruning} \textit{Upper} shows the generalization performance of the pruned models. To ease the comparison, Figure~\ref{fig:ipt_pruning} \textit{Lower} shows the change in generalization performance with respect to the un-pruned models. 

We have the following observations:

$\bullet~$ The pruning performance of the {\ours} optimized models remains higher than that of the baseline models (Figure~\ref{fig:ipt_pruning} \textit{Upper}).

$\bullet~$ Even the most redundant parameters in the {\ours} optimized models makes contributions (Figure~\ref{fig:ipt_pruning} \textit{Lower}). When there are over $80\%$ of weights remaining, the pruning performance of the baseline models is comparable or even superior than their un-pruned alternatives. In contrast, the performance of the {\ours} optimized models consistently deteriorates. This suggests that the most redundant parameters in the baseline models fail to contribute, while those in the {\ours} optimized models are trained more sufficiently and are able to make contributions. 

% $\bullet~$ Parameters contribute more equally the {\ours} optimized model in small tasks such as RTE (2.5K) (Figure~\ref{fig:ipt_pruning} \textit{Lower}). As can be seen, when there are less than $80\%$ weights remaining, the baseline performance slowly decreases until a certain ratio; then the performance drops significantly. This suggests that some parameters contribute little, while others make enormous contributions. In contrast, the performance of the {\ours} optimized model decreases almost linearly. % When there are less than $80\%$ weights remaining, both models' performances begin to degrade. Worth mentioning, in small task such as RTE (2.5K), the baseline performance slowly decreases until a certain ratio; then the performance drops significantly. This suggests that some parameters contribute little, while others make enormous contributions. In contrast, the performance of the {\ours} optimized model decreases almost linearly, suggesting all parameters contribute more uniformly. 

\begin{figure}[htb!]
    %\vspace{-0.1in}
    \centering
    \includegraphics[width=0.9\linewidth]{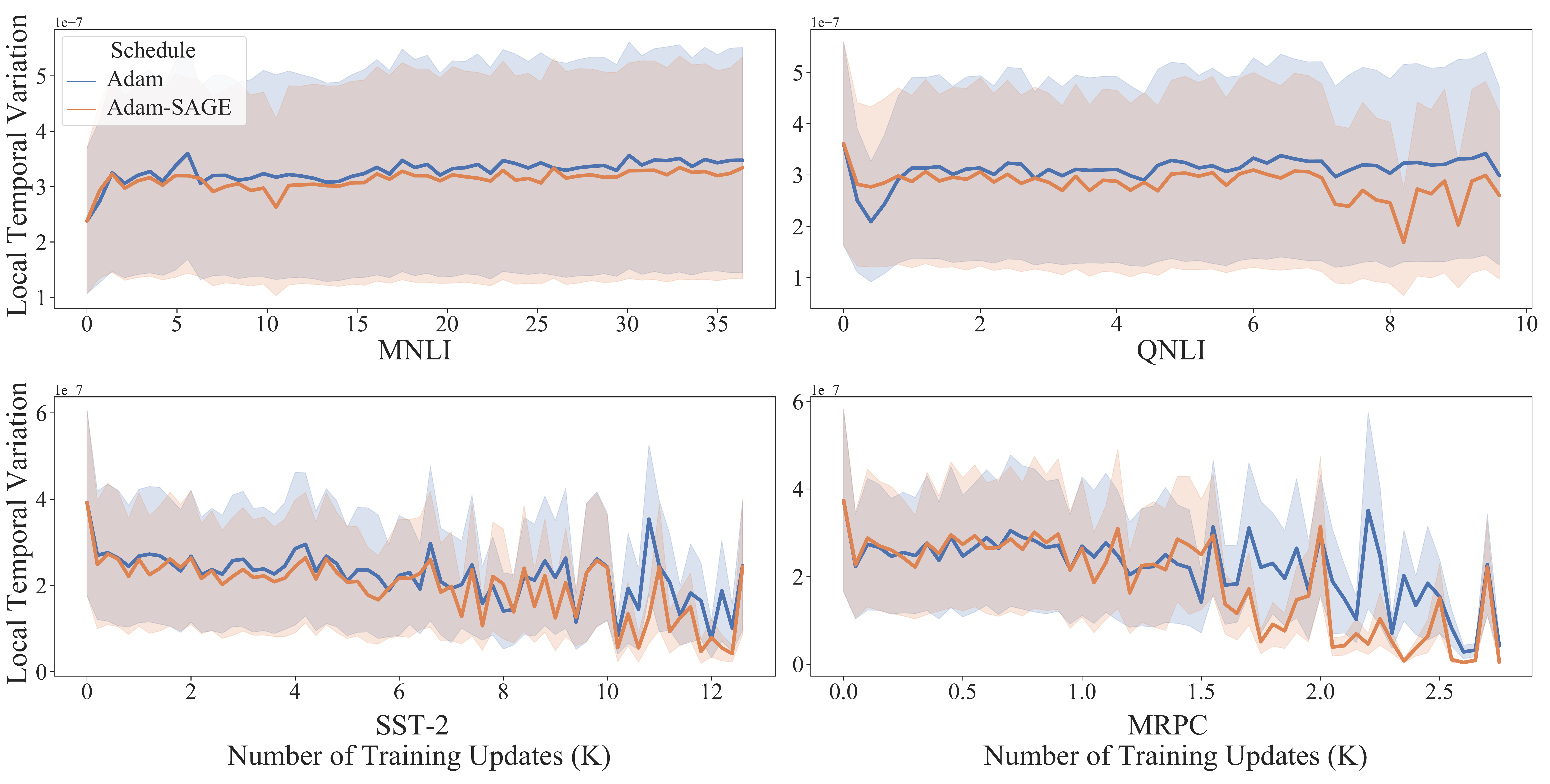}
    %\vspace{-0.05in}
	\caption{The local temporal variation of sensitivity (with $\beta_0 = 0.7$) during training.}
	\label{fig:ipt_stability}
	%\vspace{-0.15in}
\end{figure}

\textbf{Sensitivity is a reliable indicator of redundancy.} We visualize the local temporal variation (Figure~\ref{fig:ipt_stability}) to verify that sensitivity indeed becomes a more reliable indicator of redundancy in {\ours} than in the baselines. We track the variation for all parameters in the BERT-base model at each iteration, and we evaluate the variation based on the current mini-batch of training data. See Appendix~\ref{app:suff-train} for implementation details.

We observe that the local temporal variation in {\ours} remains lower or decreases faster than in the baselines for all tasks. For example, the variation in the baseline approach remains large in QNLI. In contrast, the variation in {\ours} decreases, suggesting the sensitivity indeed stabilizes and becomes a reliable indicator of redundancy.

% $\bullet~$ 
% The local temporal variation in \textit{both} the baseline and the {\ours} optimizations shows no decrement in MNLI, and shows decrement in SST-2 and MRPC. Recall that the variation measures the uncertainty in sensitivity. The uncertainty comes from not only the changes of model, but also the randomness in data sampling; while the later is not caused by optimization algorithms. In MNLI, the uncertainty from data is large and dominates the variation. As a result, the stabilization effect of {\ours} is unobvious.
%This is possibly because MNLI has a significant variance among training data. Since the variation is computed based on a randomly sampled mini-batch of the training data, the trend can be overturned by the noises. 
% In contrast, the uncertainty from data is low in SST-2 and MRPC, and thus the effect of {\ours} becomes obvious. 

\subsection{{\ours} Leads to Better Generalization Performance}
\label{sec:ana_generalization}

We verify that {\ours} leads to better generalization performance through inspecting the learning curves, decision boundary and hyper-parameter search space.

\begin{wrapfigure}{r}{0.45\textwidth}
    %\vspace{-0.2in}
    \centering
    \includegraphics[width=1\linewidth,height=0.53\linewidth]{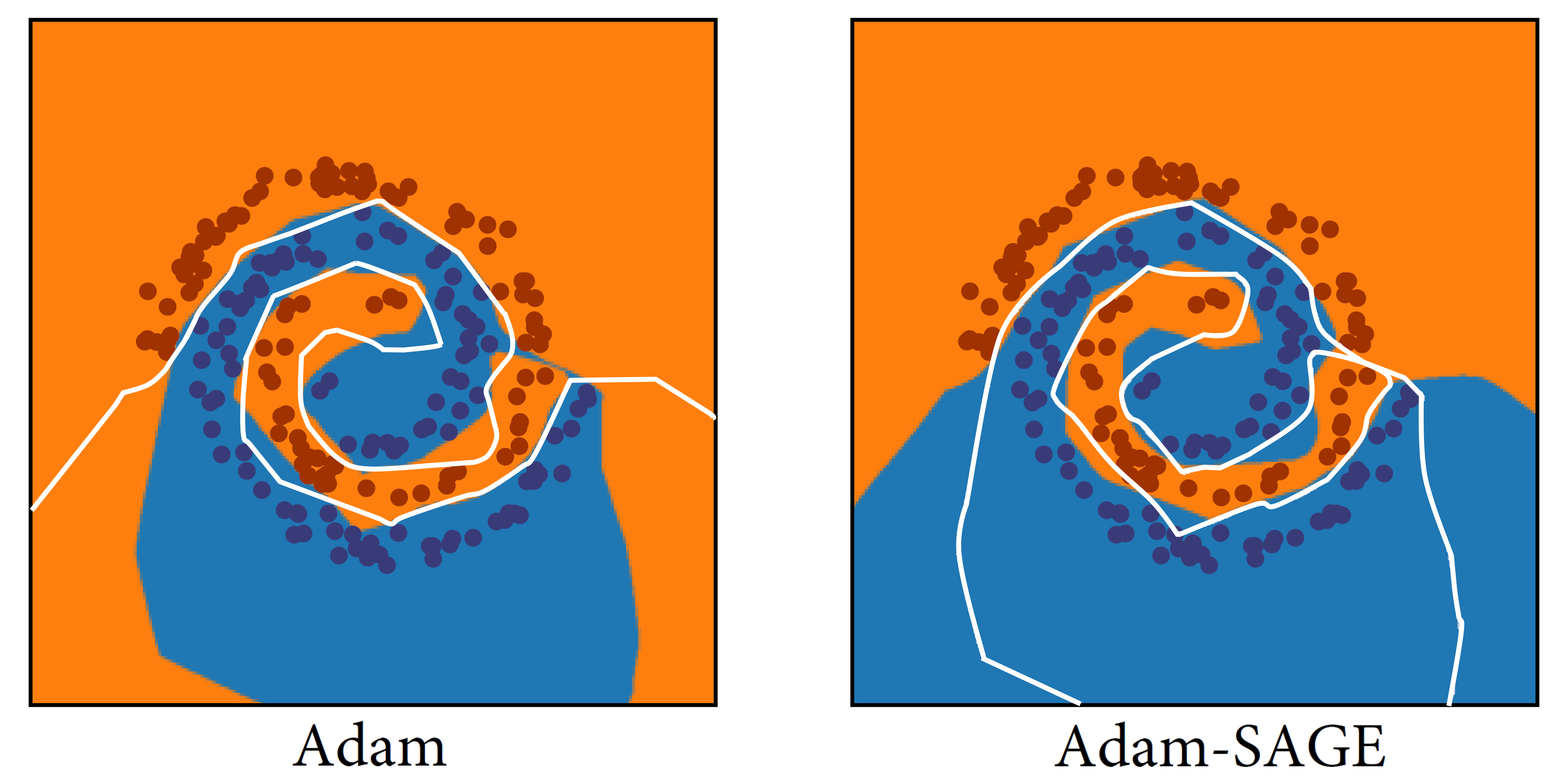}
    %\vspace{-0.3in}
	\caption{Decision boundary predicted on the Spiral dataset. The white curve on Adam-SAGE corresponds the decision boundary of Adam.}
	\label{fig:boundary}
	%\vspace{-0.2in}
\end{wrapfigure} 

\begin{figure}[htb!]
    \centering
    % \vspace{-0.05in}
    \includegraphics[width=1.0\linewidth]{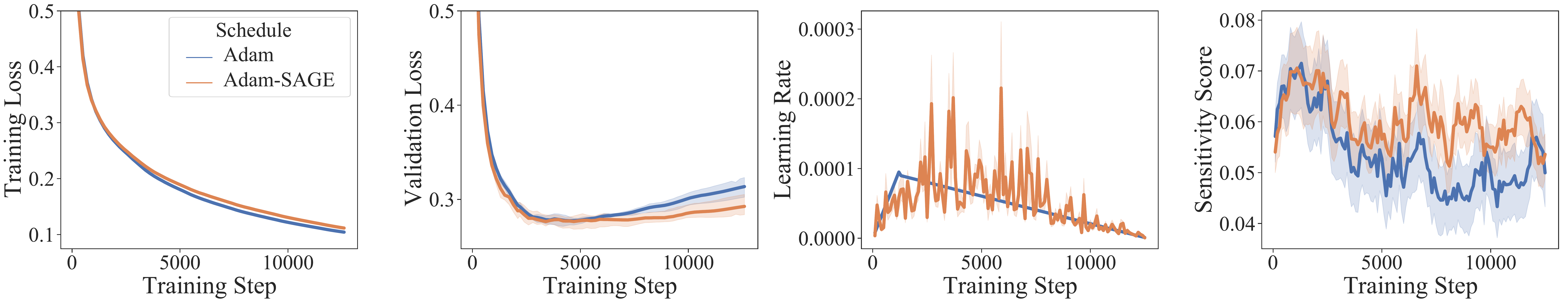}
	\caption{Learning curves obtained by fine-tuning BERT-base on SST-2 dataset.}
	\label{fig:learning_curve}
	%\vspace{-0.05in}
\end{figure}

\textbf{Learning Curves.} Figure~\ref{fig:learning_curve} shows the training loss, validation loss, learning rate, and sensitivity score obtained by fine-tuning BERT-base on SST-2. All experiment details are deferred to Appendix~\ref{app:ana_generalization}. We have two major observations: 1) {\ours}'s validation loss descends faster and {\ours} is less prone to overfitting. This observation suggests that {\ours} has a regularization effect and reduces the model variance. 2) {\ours}'s variance of the sensitivity score becomes lower through training, aligning with our observation in Figure~\ref{fig:ipt_dist}. This suggests that {\ours} gives rise to a more balanced and sufficient training. Both observations agree with our initial motivation (Figure~\ref{fig:gen_ilst}) that redundancy elimination can lead to better generalization.

\begin{figure}[htb!]
    %\vspace{-0.05in}
    \centering
    \includegraphics[width=1.0\linewidth]{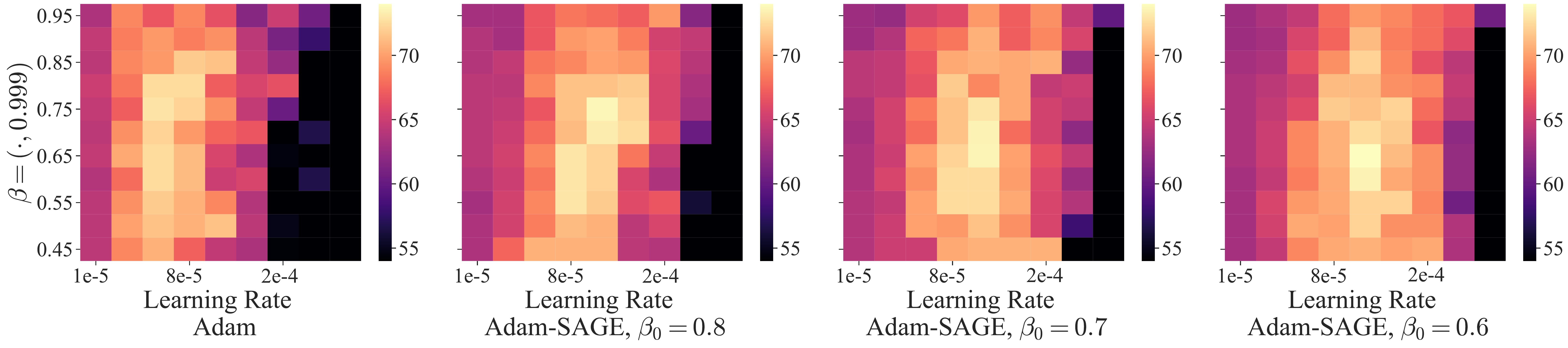}
% 	\vspace{-0.1in}
	\caption{Validation accuracy obtained by fine-tuning BERT-base on RTE dataset with a wide range of hyper-parameters.}
	\label{fig:heatmap}
	%\vspace{-0.15in}
\end{figure}

\textbf{Hyper-parameter Study.} Figure~\ref{fig:heatmap} shows the validation accuracy heatmap obtained by fine-tuning BERT-base on the RTE dataset. We plot the accuracy obtained by training with different learning rates, Adam's $\beta$s and {\ours}'s $\beta_0$s. We can observe that {\ours} consistently achieves a better generalization performance within a larger region of hyper-parameter search space under different $\beta_0$s. We also provide a hyper-parameter study for more datasets in Appendix~\ref{app:param_study}.

\textbf{Decision Boundary.} Figure~\ref{fig:boundary} shows the decision boundary predicted with Adam and {\ours} on the Spiral dataset. Specifically, we train a multi-layer perceptron with $3$ hidden layers, each with a hidden dimension of $100$. The decision boundary predicted with {\ours} is smoother and has a larger margin than with Adam, suggesting {\ours} produces a better generalized model.

%\vspace{-0.03in}
\subsection{Combine with State-Of-The-Art Methods}
\label{sec:sota_combine}
%\vspace{-0.03in}
%Both adversarial learning and {\ours} regularize sensitivity. However, there are several differences between their concepts of sensitivity:

% While adversarial learning regularizes the model's sensitivity to perturbations on \textit{input data}, {\ours} regularizes the model's sensitivity to perturbations on \textit{weight parameters}. 

% $\bullet~$ Adversarial learning aims to encourage the model's robustness to perturbations. As a result, it regularizes the model's overall sensitivity to be small.  In contrast, {\ours} aims to build a model with consistent outputs to \textit{perturbations at different places}, i.e., different weights. As a result, it regularizes sensitivity to be \textit{uniformly distributed} on all parameters. We remark that {\ours} is not designed to encourage the model's robustness to weight perturbations, but it implicitly does so. For example, as shown in Figure~\ref{fig:ipt_dist}, {\ours} lowers the overall sensitivity when the model is overfitted, and thus encourages the model's robustness. %if the original model is underfitting, {\ours} increases the overall sensitivity (e.g., Figure~\ref{fig:}) and thus discourages the model's robustness.

We further show that {\ours} is complementary to existing state-of-the-art regularization methods. Specifically, we apply {\ours} to SMART \citep{jiang2019smart}, a state-of-the-art smoothness-inducing adversarial regularization method. As shown in Table~\ref{tb:glue-smart-dev-results}, {\ours} can further improve upon SMART, suggesting the two techniques are complementary.

\newcolumntype{C}{@{\hskip3pt}c@{\hskip3pt}}
\begin{table*}[htb!]
\centering
\scriptsize
\begin{tabular}{l|l|CCCCCCCCCC}
\toprule
\multirow{2}{*}{\textbf{Model}} & \multirow{2}{*}{\textbf{Optimizer}} & \textbf{RTE} & \textbf{MRPC} & \textbf{CoLA} & \textbf{SST-2} & \textbf{STS-B} & \textbf{QNLI} & \textbf{QQP} & \textbf{MNLI-m/mm} & \textbf{Average} \\ 
&& Acc & Acc/F1 & Mcc & Acc & P/S Corr & Acc & Acc/F1 & Acc & Score \\\midrule
BERT\textsubscript{BASE} & Adamax & 69.2 & 86.2/90.4 & 57.8 & 92.9 & 89.7/89.2 & 91.2 & 90.9/88.0 & 84.5/84.4 & 82.8 \\ \midrule
\multirow{2}{*}{SMART\textsubscript{BASE}}& Adamax & 72.5 & 87.7/91.4 & 59.5 & 93.5 & 90.0/89.6 & 91.9 & 91.7/88.9 & 85.2/85.7 & 84.1 \\
& Adamax-{\ours} & \textbf{75.1} & \textbf{89.0/92.8} & \textbf{60.8} & \textbf{94.3} & \textbf{90.1/89.7} & \textbf{92.2} & \textbf{91.9/89.1} & \textbf{85.9/86.0} & \textbf{85.0} \\\bottomrule
\end{tabular}
\caption{Single task fine-tuning dev results on GLUE.}
\label{tb:glue-smart-dev-results}
\end{table*}

% \subsection{Parameter Study}
% We lastly investigate the influence of hyper-parameters learning rate and $\beta_{0}$ on the performance of {\ours} (Figure~\ref{fig:param_study}). As can be seen, {\ours} requires a larger learning rate than the baselines to offset the small scale of the modulation term (the optimal baseline learning rate lies in $5 \times 10^{-5}\sim1\times 10^{-4}$ for MNLI, $5\times 10^{-4}\sim 7\times 10^{-4}$ for IWSLT 14 De-En and $0.1\sim 0.2$ for CIFAR10). Furthermore, switching to a larger learning rate requires a lower $\beta_{0}$ to maintain the same level of performance. 
% % In all tasks, a change in $0.1$ in $\beta_{0}$ may result in a variance around $0.2$ points in accuracy, suggesting choosing a proper local window to compute the temporal variation is critical to success.

% \begin{figure}[htb!]
%     \centering
%     \vspace{-0.05in}
%     \includegraphics[width=1.0\linewidth]{Figure/parameter_study.pdf}
% 	\caption{Parameter study on learning rate and $\beta_{0}$.}
% 	\label{fig:param_study}
% 	\vspace{-0.05in}
% \end{figure}

%%%%%%%%%%%%%%%%%%%%%%%%%%%%%%%%%%%%%%%%%%%%%%%%%%%%%%%%%%%%%%%%%
%\vspace{-0.05in}
\section{Discussion}
\label{sec:disc}
%\vspace{-0.05in}

% complementary to
% ad improve efficeincy of optimizaton (landscape)
% we consider generalization 
% our exp all adam+SAGE, no conflict

\textbf{SAGE is complementary to Adaptive Gradient Methods.} Our proposed method and the mainstream adaptive gradient methods (e.g., Adam and AdaGrad) are for fundamentally different purposes. The mainstream adaptive gradient methods aim to improve optimization by adapting to the optimization landscape, while {\ours} aims to improve generalization by eliminating the weight redundancy. The quantities of our interest (i.e., Eq.~(\ref{eq:ipt}) and Eq.~(\ref{eq:std})) are related to the weight redundancy. They are not directly related to the moduli of the objective function, e.g., smoothness, curvature (which are of the interests for optimization). As shown in our experiments (See Section~\ref{sec:exp}), we do not observe any conflicts between the two methods, as {\ours} improves the model generalization performance when being combined with several adaptive gradient methods (e.g., Adam).

%\lc{Do we still need to point out "The convergence guarantees of these methods are usually established upon Lipschitz continuous gradient and bounded stochastic gradient assumptions, which do not hold for Transformer models in practice. Moreover, such convergence to the first-order stationary solutions does not imply any theoretical guarantees on the model generalization performance." here? It seems difficult to fit it in.} 

\textbf{Redundant Weights vs. Insufficiently Trained Weights.} Lottery Ticket Hypothesis \citep{frankle2018lottery} suggests that, in a randomly initialized network, there exists a well-initialized subnetwork, which outperforms any other subnetworks and matches the full model's performance. This suggests the rest parameters contribute marginally to the model performance. Although the initialization of these parameters may not be satisfactory, {\ours} provides them sufficient training so that they can learn to contribute.

% \textbf{Relation with Pruning.} Both pruning and {\ours} improve model generalization through redundancy elimination. Pruning directly zeros-out the most redundant parameters, while {\ours} reduce redundancy uniformly on all parameters. Pruning improves generalization as it can be viewed as a form of regularization to reduce model noise (Occam's razor, \citet{rasmussen2001occam}). \citet{grunwald2007minimum} further support it with a Bayesian interpretation. %Recent works provides more explanations to it all from the regularization perspective, e.g., stability enforcement, noise injection \citep{bartoldson2019generalization} and variance reduction \citep{}. 
% Different from pruning, {\ours} improves generalization as it is an explicit form of regularization on parameters with extremely high/low sensitivity, which results in a smaller function class and lower model variance.

%\vspace{-0.05in}
\section{Conclusion}
\label{con}
%\vspace{-0.05in}
We begin with a hypothesis that the redundant parameters can become useful if they are sufficiently trained by desirable optimization strategies. We verify this hypothesis by proposing an adaptive learning schedule -- {\ours}, which excites the under-fitted parameters to reduce redundancy and regularize the well-fitted parameters to prevent overfitting. We demonstrate that {\ours} can benefit model generalization in a wide range of tasks and strengthen various types of optimizers.
\bibliography{iclr2022_conference}

\begin{thebibliography}{57}
\providecommand{\natexlab}[1]{#1}
\providecommand{\url}[1]{\texttt{#1}}
\expandafter\ifx\csname urlstyle\endcsname\relax
  \providecommand{\doi}[1]{doi: #1}\else
  \providecommand{\doi}{doi: \begingroup \urlstyle{rm}\Url}\fi

\bibitem[Bar-Haim et~al.(2006)Bar-Haim, Dagan, Dolan, Ferro, and
  Giampiccolo]{rte2}
Roy Bar-Haim, Ido Dagan, Bill Dolan, Lisa Ferro, and Danilo Giampiccolo.
\newblock The second {PASCAL} recognising textual entailment challenge.
\newblock In \emph{Proceedings of the Second {PASCAL} Challenges Workshop on
  Recognising Textual Entailment}, 01 2006.

\bibitem[Bartoldson et~al.(2019)Bartoldson, Morcos, Barbu, and
  Erlebacher]{bartoldson2019generalization}
Brian~R Bartoldson, Ari~S Morcos, Adrian Barbu, and Gordon Erlebacher.
\newblock The generalization-stability tradeoff in neural network pruning.
\newblock \emph{arXiv preprint arXiv:1906.03728}, 2019.

\bibitem[Bentivogli et~al.(2009)Bentivogli, Dagan, Dang, Giampiccolo, and
  Magnini]{rte5}
Luisa Bentivogli, Ido Dagan, Hoa~Trang Dang, Danilo Giampiccolo, and Bernardo
  Magnini.
\newblock The fifth pascal recognizing textual entailment challenge.
\newblock In \emph{In Proc Text Analysis Conference (TAC’09)}, 2009.

\bibitem[Bojar et~al.(2016)Bojar, Chatterjee, Federmann, Graham, Haddow, Huck,
  Yepes, Koehn, Logacheva, Monz, et~al.]{bojar2016findings}
Ond{\v{r}}ej Bojar, Rajen Chatterjee, Christian Federmann, Yvette Graham, Barry
  Haddow, Matthias Huck, Antonio~Jimeno Yepes, Philipp Koehn, Varvara
  Logacheva, Christof Monz, et~al.
\newblock Findings of the 2016 conference on machine translation.
\newblock In \emph{Proceedings of the First Conference on Machine Translation:
  Volume 2, Shared Task Papers}, pp.\  131--198, 2016.

\bibitem[Brown et~al.(2020)Brown, Mann, Ryder, Subbiah, Kaplan, Dhariwal,
  Neelakantan, Shyam, Sastry, Askell, et~al.]{brown2020language}
Tom~B Brown, Benjamin Mann, Nick Ryder, Melanie Subbiah, Jared Kaplan, Prafulla
  Dhariwal, Arvind Neelakantan, Pranav Shyam, Girish Sastry, Amanda Askell,
  et~al.
\newblock Language models are few-shot learners.
\newblock \emph{arXiv preprint arXiv:2005.14165}, 2020.

\bibitem[Cer et~al.(2017)Cer, Diab, Agirre, Lopez-Gazpio, and
  Specia]{sts-b2017}
Daniel Cer, Mona Diab, Eneko Agirre, I{\~n}igo Lopez-Gazpio, and Lucia Specia.
\newblock Semeval-2017 task 1: Semantic textual similarity multilingual and
  crosslingual focused evaluation.
\newblock In \emph{Proceedings of the 11th International Workshop on Semantic
  Evaluation (SemEval-2017)}, pp.\  1--14, 2017.

\bibitem[Cettolo et~al.(2015)Cettolo, Niehues, St{\"u}ker, Bentivogli, Cattoni,
  and Federico]{cettolo2015iwslt}
Mauro Cettolo, Jan Niehues, Sebastian St{\"u}ker, Luisa Bentivogli, Roldano
  Cattoni, and Marcello Federico.
\newblock The iwslt 2015 evaluation campaign.
\newblock In \emph{IWSLT 2015, International Workshop on Spoken Language
  Translation}, 2015.

\bibitem[Chen et~al.(2020)Chen, Frankle, Chang, Liu, Zhang, Wang, and
  Carbin]{chen2020lottery}
Tianlong Chen, Jonathan Frankle, Shiyu Chang, Sijia Liu, Yang Zhang, Zhangyang
  Wang, and Michael Carbin.
\newblock The lottery ticket hypothesis for pre-trained bert networks.
\newblock \emph{arXiv preprint arXiv:2007.12223}, 2020.

\bibitem[Dagan et~al.(2006)Dagan, Glickman, and Magnini]{rte1}
Ido Dagan, Oren Glickman, and Bernardo Magnini.
\newblock The pascal recognising textual entailment challenge.
\newblock In \emph{Proceedings of the First International Conference on Machine
  Learning Challenges: Evaluating Predictive Uncertainty Visual Object
  Classification, and Recognizing Textual Entailment}, MLCW'05, pp.\  177--190,
  Berlin, Heidelberg, 2006. Springer-Verlag.
\newblock ISBN 3-540-33427-0, 978-3-540-33427-9.
\newblock \doi{10.1007/11736790_9}.
\newblock URL \url{http://dx.doi.org/10.1007/11736790_9}.

\bibitem[Deng et~al.(2009)Deng, Dong, Socher, Li, Li, and
  Fei-Fei]{deng2009imagenet}
Jia Deng, Wei Dong, Richard Socher, Li-Jia Li, Kai Li, and Li~Fei-Fei.
\newblock Imagenet: A large-scale hierarchical image database.
\newblock In \emph{2009 IEEE conference on computer vision and pattern
  recognition}, pp.\  248--255. Ieee, 2009.

\bibitem[Devlin et~al.(2018)Devlin, Chang, Lee, and Toutanova]{devlin2018bert}
Jacob Devlin, Ming-Wei Chang, Kenton Lee, and Kristina Toutanova.
\newblock Bert: Pre-training of deep bidirectional transformers for language
  understanding.
\newblock \emph{arXiv preprint arXiv:1810.04805}, 2018.

\bibitem[Ding et~al.(2019)Ding, Ding, Zhou, Guo, Han, and Liu]{ding2019global}
Xiaohan Ding, Guiguang Ding, Xiangxin Zhou, Yuchen Guo, Jungong Han, and
  Ji~Liu.
\newblock Global sparse momentum sgd for pruning very deep neural networks.
\newblock \emph{arXiv preprint arXiv:1909.12778}, 2019.

\bibitem[Dolan \& Brockett(2005)Dolan and Brockett]{mrpc2005}
William~B Dolan and Chris Brockett.
\newblock Automatically constructing a corpus of sentential paraphrases.
\newblock In \emph{Proceedings of the Third International Workshop on
  Paraphrasing (IWP2005)}, 2005.

\bibitem[Dosovitskiy et~al.(2020)Dosovitskiy, Beyer, Kolesnikov, Weissenborn,
  Zhai, Unterthiner, Dehghani, Minderer, Heigold, Gelly,
  et~al.]{dosovitskiy2020image}
Alexey Dosovitskiy, Lucas Beyer, Alexander Kolesnikov, Dirk Weissenborn,
  Xiaohua Zhai, Thomas Unterthiner, Mostafa Dehghani, Matthias Minderer, Georg
  Heigold, Sylvain Gelly, et~al.
\newblock An image is worth 16x16 words: Transformers for image recognition at
  scale.
\newblock \emph{arXiv preprint arXiv:2010.11929}, 2020.

\bibitem[Duchi et~al.(2011)Duchi, Hazan, and Singer]{duchi2011adaptive}
John Duchi, Elad Hazan, and Yoram Singer.
\newblock Adaptive subgradient methods for online learning and stochastic
  optimization.
\newblock \emph{Journal of machine learning research}, 12\penalty0 (7), 2011.

\bibitem[Fan et~al.(2019)Fan, Grave, and Joulin]{fan2019reducing}
Angela Fan, Edouard Grave, and Armand Joulin.
\newblock Reducing transformer depth on demand with structured dropout.
\newblock \emph{arXiv preprint arXiv:1909.11556}, 2019.

\bibitem[Foret et~al.(2020)Foret, Kleiner, Mobahi, and
  Neyshabur]{foret2020sharpness}
Pierre Foret, Ariel Kleiner, Hossein Mobahi, and Behnam Neyshabur.
\newblock Sharpness-aware minimization for efficiently improving
  generalization.
\newblock \emph{arXiv preprint arXiv:2010.01412}, 2020.

\bibitem[Frankle \& Carbin(2018)Frankle and Carbin]{frankle2018lottery}
Jonathan Frankle and Michael Carbin.
\newblock The lottery ticket hypothesis: Finding sparse, trainable neural
  networks.
\newblock \emph{arXiv preprint arXiv:1803.03635}, 2018.

\bibitem[Giampiccolo et~al.(2007)Giampiccolo, Magnini, Dagan, and Dolan]{rte3}
Danilo Giampiccolo, Bernardo Magnini, Ido Dagan, and Bill Dolan.
\newblock The third {PASCAL} recognizing textual entailment challenge.
\newblock In \emph{Proceedings of the {ACL}-{PASCAL} Workshop on Textual
  Entailment and Paraphrasing}, pp.\  1--9, Prague, June 2007. Association for
  Computational Linguistics.
\newblock URL \url{https://www.aclweb.org/anthology/W07-1401}.

\bibitem[Gr{\"u}nwald \& Grunwald(2007)Gr{\"u}nwald and
  Grunwald]{grunwald2007minimum}
Peter~D Gr{\"u}nwald and Abhijit Grunwald.
\newblock \emph{The minimum description length principle}.
\newblock 2007.

\bibitem[He et~al.(2020)He, Liu, Gao, and Chen]{he2020deberta}
Pengcheng He, Xiaodong Liu, Jianfeng Gao, and Weizhu Chen.
\newblock Deberta: Decoding-enhanced bert with disentangled attention.
\newblock \emph{arXiv preprint arXiv:2006.03654}, 2020.

\bibitem[Hinton et~al.(2012)Hinton, Srivastava, and Swersky]{hinton2012neural}
Geoffrey Hinton, Nitish Srivastava, and Kevin Swersky.
\newblock Neural networks for machine learning lecture 6a overview of
  mini-batch gradient descent.
\newblock \emph{Cited on}, 14\penalty0 (8), 2012.

\bibitem[Hou et~al.(2020)Hou, Huang, Shang, Jiang, Chen, and
  Liu]{hou2020dynabert}
Lu~Hou, Zhiqi Huang, Lifeng Shang, Xin Jiang, Xiao Chen, and Qun Liu.
\newblock Dynabert: Dynamic bert with adaptive width and depth.
\newblock \emph{arXiv preprint arXiv:2004.04037}, 2020.

\bibitem[Jiang et~al.(2019)Jiang, He, Chen, Liu, Gao, and Zhao]{jiang2019smart}
Haoming Jiang, Pengcheng He, Weizhu Chen, Xiaodong Liu, Jianfeng Gao, and Tuo
  Zhao.
\newblock Smart: Robust and efficient fine-tuning for pre-trained natural
  language models through principled regularized optimization.
\newblock \emph{arXiv preprint arXiv:1911.03437}, 2019.

\bibitem[Kingma \& Ba(2014)Kingma and Ba]{kingma2014adam}
Diederik~P Kingma and Jimmy Ba.
\newblock Adam: A method for stochastic optimization.
\newblock \emph{arXiv preprint arXiv:1412.6980}, 2014.

\bibitem[Krizhevsky et~al.(2009)Krizhevsky, Hinton,
  et~al.]{krizhevsky2009learning}
Alex Krizhevsky, Geoffrey Hinton, et~al.
\newblock Learning multiple layers of features from tiny images.
\newblock 2009.

\bibitem[LeCun et~al.(1990)LeCun, Denker, and Solla]{lecun1990optimal}
Yann LeCun, John~S Denker, and Sara~A Solla.
\newblock Optimal brain damage.
\newblock In \emph{Advances in neural information processing systems}, pp.\
  598--605, 1990.

\bibitem[Lee et~al.(2018)Lee, Ajanthan, and Torr]{lee2018snip}
Namhoon Lee, Thalaiyasingam Ajanthan, and Philip~HS Torr.
\newblock Snip: Single-shot network pruning based on connection sensitivity.
\newblock \emph{arXiv preprint arXiv:1810.02340}, 2018.

\bibitem[Liang et~al.(2021)Liang, Zuo, Chen, Jiang, Liu, He, Zhao, and
  Chen]{liang2021super}
Chen Liang, Simiao Zuo, Minshuo Chen, Haoming Jiang, Xiaodong Liu, Pengcheng
  He, Tuo Zhao, and Weizhu Chen.
\newblock Super tickets in pre-trained language models: From model compression
  to improving generalization.
\newblock \emph{arXiv preprint arXiv:2105.12002}, 2021.

\bibitem[Liu et~al.(2019{\natexlab{a}})Liu, Jiang, He, Chen, Liu, Gao, and
  Han]{liu2019variance}
Liyuan Liu, Haoming Jiang, Pengcheng He, Weizhu Chen, Xiaodong Liu, Jianfeng
  Gao, and Jiawei Han.
\newblock On the variance of the adaptive learning rate and beyond.
\newblock \emph{arXiv preprint arXiv:1908.03265}, 2019{\natexlab{a}}.

\bibitem[Liu et~al.(2020)Liu, Wang, Ji, Cheng, Zhu, Awa, He, Chen, Poon, Cao,
  et~al.]{liu2020mtmtdnn}
Xiaodong Liu, Yu~Wang, Jianshu Ji, Hao Cheng, Xueyun Zhu, Emmanuel Awa,
  Pengcheng He, Weizhu Chen, Hoifung Poon, Guihong Cao, et~al.
\newblock The microsoft toolkit of multi-task deep neural networks for natural
  language understanding.
\newblock In \emph{Proceedings of the 58th Annual Meeting of the Association
  for Computational Linguistics: System Demonstrations}, pp.\  118--126, 2020.

\bibitem[Liu et~al.(2019{\natexlab{b}})Liu, Ott, Goyal, Du, Joshi, Chen, Levy,
  Lewis, Zettlemoyer, and Stoyanov]{liu2019roberta}
Yinhan Liu, Myle Ott, Naman Goyal, Jingfei Du, Mandar Joshi, Danqi Chen, Omer
  Levy, Mike Lewis, Luke Zettlemoyer, and Veselin Stoyanov.
\newblock Roberta: A robustly optimized bert pretraining approach.
\newblock \emph{arXiv preprint arXiv:1907.11692}, 2019{\natexlab{b}}.

\bibitem[Liu et~al.(2018)Liu, Sun, Zhou, Huang, and Darrell]{liu2018rethinking}
Zhuang Liu, Mingjie Sun, Tinghui Zhou, Gao Huang, and Trevor Darrell.
\newblock Rethinking the value of network pruning.
\newblock \emph{arXiv preprint arXiv:1810.05270}, 2018.

\bibitem[Loshchilov \& Hutter(2018)Loshchilov and Hutter]{loshchilov2018fixing}
Ilya Loshchilov and Frank Hutter.
\newblock Fixing weight decay regularization in adam.
\newblock 2018.

\bibitem[Lubana \& Dick(2020)Lubana and Dick]{lubana2020gradient}
Ekdeep~Singh Lubana and Robert~P Dick.
\newblock A gradient flow framework for analyzing network pruning.
\newblock \emph{arXiv preprint arXiv:2009.11839}, 2020.

\bibitem[Michel et~al.(2019)Michel, Levy, and Neubig]{michel2019sixteen}
Paul Michel, Omer Levy, and Graham Neubig.
\newblock Are sixteen heads really better than one?
\newblock \emph{arXiv preprint arXiv:1905.10650}, 2019.

\bibitem[Molchanov et~al.(2016)Molchanov, Tyree, Karras, Aila, and
  Kautz]{molchanov2016pruning}
Pavlo Molchanov, Stephen Tyree, Tero Karras, Timo Aila, and Jan Kautz.
\newblock Pruning convolutional neural networks for resource efficient
  inference.
\newblock \emph{arXiv preprint arXiv:1611.06440}, 2016.

\bibitem[Molchanov et~al.(2019)Molchanov, Mallya, Tyree, Frosio, and
  Kautz]{molchanov2019importance}
Pavlo Molchanov, Arun Mallya, Stephen Tyree, Iuri Frosio, and Jan Kautz.
\newblock Importance estimation for neural network pruning.
\newblock In \emph{Proceedings of the IEEE/CVF Conference on Computer Vision
  and Pattern Recognition}, pp.\  11264--11272, 2019.

\bibitem[Morcos et~al.(2019)Morcos, Yu, Paganini, and Tian]{morcos2019one}
Ari~S Morcos, Haonan Yu, Michela Paganini, and Yuandong Tian.
\newblock One ticket to win them all: generalizing lottery ticket
  initializations across datasets and optimizers.
\newblock \emph{arXiv preprint arXiv:1906.02773}, 2019.

\bibitem[Mozer \& Smolensky(1989)Mozer and Smolensky]{mozer1989skeletonization}
Michael~C Mozer and Paul Smolensky.
\newblock Skeletonization: A technique for trimming the fat from a network via
  relevance assessment.
\newblock In \emph{Advances in neural information processing systems}, pp.\
  107--115, 1989.

\bibitem[Ott et~al.(2018)Ott, Edunov, Grangier, and Auli]{ott2018scaling}
Myle Ott, Sergey Edunov, David Grangier, and Michael Auli.
\newblock Scaling neural machine translation.
\newblock \emph{arXiv preprint arXiv:1806.00187}, 2018.

\bibitem[Ott et~al.(2019)Ott, Edunov, Baevski, Fan, Gross, Ng, Grangier, and
  Auli]{ott2019fairseq}
Myle Ott, Sergey Edunov, Alexei Baevski, Angela Fan, Sam Gross, Nathan Ng,
  David Grangier, and Michael Auli.
\newblock fairseq: A fast, extensible toolkit for sequence modeling.
\newblock In \emph{Proceedings of NAACL-HLT 2019: Demonstrations}, 2019.

\bibitem[Rajpurkar et~al.(2016)Rajpurkar, Zhang, Lopyrev, and Liang]{squad1}
Pranav Rajpurkar, Jian Zhang, Konstantin Lopyrev, and Percy Liang.
\newblock {SQ}u{AD}: 100,000+ questions for machine comprehension of text.
\newblock In \emph{Proceedings of the 2016 Conference on Empirical Methods in
  Natural Language Processing}, pp.\  2383--2392, Austin, Texas, November 2016.
  Association for Computational Linguistics.
\newblock \doi{10.18653/v1/D16-1264}.
\newblock URL \url{https://www.aclweb.org/anthology/D16-1264}.

\bibitem[Rasmussen \& Ghahramani(2001)Rasmussen and
  Ghahramani]{rasmussen2001occam}
Carl~Edward Rasmussen and Zoubin Ghahramani.
\newblock Occam's razor.
\newblock \emph{Advances in neural information processing systems}, pp.\
  294--300, 2001.

\bibitem[Robbins \& Monro(1951)Robbins and Monro]{robbins1951stochastic}
Herbert Robbins and Sutton Monro.
\newblock A stochastic approximation method.
\newblock \emph{The annals of mathematical statistics}, pp.\  400--407, 1951.

\bibitem[Sanh et~al.(2020)Sanh, Wolf, and Rush]{sanh2020movement}
Victor Sanh, Thomas Wolf, and Alexander~M Rush.
\newblock Movement pruning: Adaptive sparsity by fine-tuning.
\newblock \emph{arXiv preprint arXiv:2005.07683}, 2020.

\bibitem[Sennrich et~al.(2016)Sennrich, Haddow, and
  Birch]{sennrich2016edinburgh}
Rico Sennrich, Barry Haddow, and Alexandra Birch.
\newblock Edinburgh neural machine translation systems for wmt 16.
\newblock \emph{arXiv preprint arXiv:1606.02891}, 2016.

\bibitem[Socher et~al.(2013)Socher, Perelygin, Wu, Chuang, Manning, Ng, and
  Potts]{sst2013}
Richard Socher, Alex Perelygin, Jean Wu, Jason Chuang, Christopher~D Manning,
  Andrew Ng, and Christopher Potts.
\newblock Recursive deep models for semantic compositionality over a sentiment
  treebank.
\newblock In \emph{Proceedings of the 2013 conference on empirical methods in
  natural language processing}, pp.\  1631--1642, 2013.

\bibitem[Theis et~al.(2018)Theis, Korshunova, Tejani, and
  Husz{\'a}r]{theis2018faster}
Lucas Theis, Iryna Korshunova, Alykhan Tejani, and Ferenc Husz{\'a}r.
\newblock Faster gaze prediction with dense networks and fisher pruning.
\newblock \emph{arXiv preprint arXiv:1801.05787}, 2018.

\bibitem[Vaswani et~al.(2017)Vaswani, Shazeer, Parmar, Uszkoreit, Jones, Gomez,
  Kaiser, and Polosukhin]{vaswani2017attention}
Ashish Vaswani, Noam Shazeer, Niki Parmar, Jakob Uszkoreit, Llion Jones,
  Aidan~N Gomez, Lukasz Kaiser, and Illia Polosukhin.
\newblock Attention is all you need.
\newblock \emph{arXiv preprint arXiv:1706.03762}, 2017.

\bibitem[Voita et~al.(2019)Voita, Talbot, Moiseev, Sennrich, and
  Titov]{voita2019analyzing}
Elena Voita, David Talbot, Fedor Moiseev, Rico Sennrich, and Ivan Titov.
\newblock Analyzing multi-head self-attention: Specialized heads do the heavy
  lifting, the rest can be pruned.
\newblock \emph{arXiv preprint arXiv:1905.09418}, 2019.

\bibitem[Wang et~al.(2018)Wang, Singh, Michael, Hill, Levy, and
  Bowman]{wang2018glue}
Alex Wang, Amanpreet Singh, Julian Michael, Felix Hill, Omer Levy, and Samuel~R
  Bowman.
\newblock Glue: A multi-task benchmark and analysis platform for natural
  language understanding.
\newblock \emph{arXiv preprint arXiv:1804.07461}, 2018.

\bibitem[Wang et~al.(2019)Wang, Wohlwend, and Lei]{wang2019structured}
Ziheng Wang, Jeremy Wohlwend, and Tao Lei.
\newblock Structured pruning of large language models.
\newblock \emph{arXiv preprint arXiv:1910.04732}, 2019.

\bibitem[Warstadt et~al.(2019)Warstadt, Singh, and Bowman]{cola2018}
Alex Warstadt, Amanpreet Singh, and Samuel~R Bowman.
\newblock Neural network acceptability judgments.
\newblock \emph{Transactions of the Association for Computational Linguistics},
  7:\penalty0 625--641, 2019.

\bibitem[Williams et~al.(2018)Williams, Nangia, and Bowman]{mnli2018}
Adina Williams, Nikita Nangia, and Samuel Bowman.
\newblock A broad-coverage challenge corpus for sentence understanding through
  inference.
\newblock In \emph{Proceedings of the 2018 Conference of the North American
  Chapter of the Association for Computational Linguistics: Human Language
  Technologies, Volume 1 (Long Papers)}, pp.\  1112--1122. Association for
  Computational Linguistics, 2018.
\newblock URL \url{http://aclweb.org/anthology/N18-1101}.

\bibitem[Xiao et~al.(2019)Xiao, Wang, and Rajasekaran]{xiao2019autoprune}
Xia Xiao, Zigeng Wang, and Sanguthevar Rajasekaran.
\newblock Autoprune: Automatic network pruning by regularizing auxiliary
  parameters.
\newblock \emph{Advances in neural information processing systems}, 32, 2019.

\bibitem[Zeiler(2012)]{zeiler2012adadelta}
Matthew~D Zeiler.
\newblock Adadelta: an adaptive learning rate method.
\newblock \emph{arXiv preprint arXiv:1212.5701}, 2012.

\end{thebibliography}
\bibliographystyle{iclr2022_conference}
\clearpage
\appendix
%!TEX root = main.tex
\clearpage
\section{Appendix}
\label{appendix}

\subsection{Natural Language Understanding}

\subsubsection{Data}
\label{app:glue-data}

GLUE is a collection of nine NLU tasks. The benchmark includes question answering \citep{squad1}, linguistic acceptability (CoLA, \citealt{cola2018}), sentiment analysis (SST, \citealt{sst2013}), text similarity (STS-B, \citealt{sts-b2017}), paraphrase detection (MRPC, \citealt{mrpc2005}), and natural language inference (RTE \& MNLI, \citealt{rte1,rte2,rte3,rte5,mnli2018}) tasks. Details of the GLUE benchmark, including tasks, statistics, and evaluation metrics, are summarized in Table~\ref{tab:glue}. 

All the texts were tokenized using wordpieces, and were chopped to spans no longer than $512$ tokens. 

\subsubsection{Training Details}
\label{app:glue-training}

To fine-tune BERT-base and RoBERTa-large models on individual tasks, we append a task-specific fully-connected classification layer to them as in \citet{devlin2018bert}.

Table~\ref{tb:glue-hyperparam} present the hyper-parameter configurations. We tune this set of hyper-parameters on a single seed, and report the averaged results obtained with the same configuration over all seeds. For {\ours} experiments, We slightly tune $\beta_{0}$ within a range of $0.1$ on different seeds. We apply a linear weight decay rate of $0.01$ and a gradient norm clipping threshold of $1$ for all experiments. All experiments are conducted on Nvidia V100 GPUs.

\newcolumntype{C}{@{\hskip3pt}c@{\hskip3pt}}
\begin{table*}[htb!]
\centering
\footnotesize
\begin{tabular}{l|l|CCCCCCCCC}
\toprule
\textbf{Hyper-param} & \textbf{Experiment}                      &\textbf{RTE} & \textbf{MRPC} & \textbf{CoLA} & \textbf{SST-2} & \textbf{STS-B} & \textbf{QNLI} & \textbf{QQP} & \textbf{MNLI} \\ \midrule
\multirow{6}{*}{Learning Rate} & BERT\textsubscript{BASE}, Adam & 1e-5 & 1e-5 & 1e-5 & 1e-5 & 1e-5 & 1e-5 & 2e-5 & 2e-5 \\
& BERT\textsubscript{BASE}, Adam-{\ours}                        & 1e-4 & 8e-5 & 8e-5 & 3e-5 & 1e-4 & 8e-5 & 4e-5 & 5e-5 \\
& BERT\textsubscript{BASE}, Adamax & 1e-4 & 1e-4 & 1e-4 & 5e-5 & 1e-4 & 1e-4 & 1e-4 & 8e-5 \\
& BERT\textsubscript{BASE}, Adamax-{\ours} & 3e-4 & 3e-4 & 2e-4 & 2e-4 & 5e-4 & 5e-4 & 3e-4 &  2e-4\\
& RoBERTa\textsubscript{LARGE}, Adamax & 5e-5 & 5e-5 & 3e-5 & 1e-5 & 5e-5 & 1e-5 & 1e-4 & 1e-5 \\
& RoBERTa\textsubscript{LARGE}, Adamax-{\ours} & 6e-5 & 2e-4 & 8e-5 & 2e-5 & 8e-5 & 3e-5 & 2e-4 & 8e-5 \\\midrule
\multirow{3}{*}{$\beta_{0}$} & BERT\textsubscript{BASE}, Adam-{\ours} & 0.60 & 0.80 & 0.70 & 0.80 & 0.60 & 0.70 & 0.75 & 0.70 \\
& BERT\textsubscript{BASE}, Adamax-{\ours} & 0.65 & 0.80 & 0.75 & 0.70 & 0.75 & 0.70 & 0.75 & 0.85 \\
& RoBERTa\textsubscript{LARGE}, Adamax-{\ours} & 0.75 & 0.65 & 0.70 & 0.75 & 0.80 & 0.80 & 0.65 & 0.60 \\\midrule
\multirow{2}{*}{Batch Size} & BERT\textsubscript{BASE} & 16 & 8 & 32 & 32 & 32 & 32 & 32 & 32\\
& RoBERTa\textsubscript{LARGE} & 16 & 8 & 32 & 32 & 32 & 32 & 32 & 32\\\midrule
\multirow{2}{*}{Epoch} & BERT\textsubscript{BASE} & 6 & 6 & 6 & 6 & 6 & 3 & 6 & 3 \\
& RoBERTa\textsubscript{LARGE} & 15 & 6 & 6 & 6 & 10 & 10 & 15 & 3  \\\midrule
\multirow{2}{*}{Dropout} & BERT\textsubscript{BASE} & 0.1 & 0.1 & 0.1 & 0.1 & 0.1 & 0.1 & 0.0 & 0.3 \\
& RoBERTa\textsubscript{LARGE} & 0.1 & 0.1 & 0.1 & 0.1 & 0.1 & 0.1 & 0.0 & 0.3 \\\midrule
\multirow{2}{*}{Warmup} & BERT\textsubscript{BASE} & 0.1 & 0.1 & 0.1 & 0.1 & 0.1 & 0.1 & 0.0 & 0.1  \\
& RoBERTa\textsubscript{LARGE} & 0.1 & 0.1 & 0.1 & 0.1 & 0.1 & 0.1 & 0.1 & 0.1  \\\bottomrule
\end{tabular}
\caption{Hyper-parameter configurations for GLUE experiments. ``Epoch'' refers to the total training epochs; we adopt early-stopping strategy in practice. ``Dropout'' refers to classification layer dropout ratio. ``Warmup'' refers to the ratio of learning rate linear warmup iterations to total training iterations. }
\label{tb:glue-hyperparam}
\end{table*}

\subsubsection{Evaluation Results}
\label{app:glue-exp}
\textbf{Statistics of the dev set results.} Table~\ref{tb:glue-dev-stats} shows the standard deviation of the dev set results.
\newcolumntype{C}{@{\hskip3pt}c@{\hskip3pt}}
\begin{table*}[htb!]
\centering
\footnotesize
\begin{tabular}{l|l|CCCCCCCCC}
\toprule
\textbf{Model} & \textbf{Optimizer}  & \textbf{RTE} & \textbf{MRPC} & \textbf{CoLA} & \textbf{SST-2} & \textbf{STS-B} & \textbf{QNLI} & \textbf{QQP} & \textbf{MNLI} \\ \midrule
\multirow{2}{*}{BERT\textsubscript{BASE}} & Adam-{\ours} & 0.35 & 0.32 & 0.85 & 0.25 & 0.12 & 0.06 & 0.05 & 0.06 \\ \cmidrule(r){2-11}
%& Adamax & 1.37 & 0.55 & 0.35 & 0.20 & 0.06 & 0.10 & 0.09 & 0.21 \\
& Adamax-{\ours} & 0.56 & 0.69 & 0.12 & 0.23 & 0.03 & 0.06 & 0.08 & 0.10 \\ \midrule
RoBERTa\textsubscript{LARGE} %& Adamax & 1.43 & 0.92 & 0.50 & 0.06 & 0.11 & 0.10 & 0.10 & 0.25 &  \\
& Adamax-{\ours} & 0.51 & 0.78 & 0.50 & 0.19 & 0.08 & 0.00 & 0.05 & 0.05 &\\ \bottomrule
\end{tabular}
\caption{Standard deviation of the dev set results.}
\label{tb:glue-dev-stats}
\end{table*}

\textbf{Average score computation formula.} For dev set results, we first obtain a score for each task by averaging the scores of all metrics (e.g., Acc and F1) and test sets (e.g., MNLI-m and MNLI-mm) within this task, then compute a task-average score. For test set results, we directly averages scores of all reported metrics following \citet{devlin2018bert}.

\subsection{Neural Machine Translation}

\subsubsection{Data}
\label{app:nmt-data}

Table~\ref{tab:nmt-data} shows the number of sentence pairs in each dataset. We use the standard newstest-2013 and newstest-2014 as dev and test set for WMT'16 En-De. We follow \citet{ott2019fairseq} to split the dev/test sets for IWSLT'14 De-En. 

All datasets are encoded using byte-pair encoding (BPE, \citet{sennrich2016edinburgh}). We preprocess IWSLT'14 De-En data following \textit{fairseq}\footnote{https://github.com/pytorch/fairseq/blob/master/examples/translation} and adopt the preprocessed WMT'16 En-De from Google\footnote{https://pytorchnlp.readthedocs.io/en/latest/\_modules/torchnlp/datasets/wmt.html}.

\newcolumntype{C}{@{\hskip3pt}c@{\hskip3pt}}
\begin{table*}[!htb]
\centering
\begin{tabular}{l|CCC} \toprule 
\textbf{Data}  &  \textbf{Train} & \textbf{Dev} & \textbf{Test} \\ \midrule
\textbf{IWSLT'14 De-En} &  160K & 7283  & 6750 \\
\textbf{WMT'16 En-De} &  4.5M & 1061  & 1019 \\\bottomrule
\end{tabular}
\caption{The number of parallel sentences in NMT datasets.}
\label{tab:nmt-data}
\end{table*} 

\subsubsection{Training Details}
\label{app:nmt-training}

We adopt the Transformer-base model for both datasets. For IWSLT'14 De-En, we share the decoder and encoder output embeddings. For WMT'16 En-De, we share all the embeddings.

Table~\ref{tb:nmt-hyperparam} presents the hyper-parameter configurations for the best models. We apply a linear weight decay rate of $1\times10^{-4}$ and a label smoothing ratio of $0.1$ for all experiments. All experiments are conducted on Nvidia V100 GPUs.

For IWSLT'14 De-En, we report the BLEU score of the best checkpoint using a beam size of $5$ and length penalty of $1$. For WMT'16 En-De, we report the average of the last $10$ checkpoints with a beam size of $4$ and length penalty of $0.6$.

\newcolumntype{C}{@{\hskip3pt}c@{\hskip3pt}}
\begin{table*}[htb!]
\centering
\small
\begin{tabular}{l|l|CC}
\toprule
\textbf{Hyper-param} & \textbf{Experiment} & \textbf{IWSLT'14 De-En} & \textbf{WMT'16 En-De} \\ \midrule
\multirow{2}{*}{Learning Rate} & Adam & 5e-4 & 7e-4 \\
& Adam-{\ours} & 1e-3 & 2e-3 \\\midrule
\multirow{1}{*}{$\beta_{0}$} & Adam-{\ours} & 0.8 & 0.4  \\\midrule
\multirow{1}{*}{Batch size} & Both & 4096 & 32768 \\\midrule
\multirow{1}{*}{Epoch} & Both & 60 & 40  \\\midrule
\multirow{1}{*}{Dropout} & Both & 0.3 & 0.1  \\\midrule
\multirow{1}{*}{Warmup} & Both & 8000 & 4000 \\\bottomrule
\end{tabular}
\caption{Hyper-parameter configurations for NMT experiments. ``Warmup'' refers to the learning rate linear warmup iterations.}
\label{tb:nmt-hyperparam}
\end{table*}

\subsection{Image Classification}

\subsubsection{Data}
\label{app:img-data}
For CIFAR100, we apply random cropping and random horizontal flipping to the training data. 

\subsubsection{Training Details}
\label{app:img-training}

Table~\ref{tb:vit-hyperparam} present the hyper-parameter configurations for the best models. All experiments are conducted on Nvidia V100 GPUs. 

\newcolumntype{C}{@{\hskip3pt}c@{\hskip3pt}}
\begin{table*}[htb!]
\centering
\small
\begin{tabular}{l|l|CC}
\toprule
\textbf{Hyper-param} & \textbf{Experiment} & \textbf{CIFAR100} & \textbf{ImageNet} \\ \midrule
\multirow{2}{*}{Learning Rate} & ViT-B/32, SGD-{\ours} & 0.02 & 0.05 \\
& ViT-L/32, SGD-{\ours} & 0.02 & 0.08 \\
\midrule
\multirow{2}{*}{$\beta_{0}$} & ViT-B/32, SGD-{\ours} & 0.95 & 0.95 \\
& ViT-L/32, SGD-{\ours} & 0.85 & 0.95 \\
\midrule
\multirow{1}{*}{Training Steps} & All & 10000 & 20000 \\
\midrule
\multirow{1}{*}{Dropout} & All & 0.0 & 0.0 \\
\bottomrule
\end{tabular}
\caption{Hyper-parameter configurations for ViT experiments on CIFAR100 and ImageNet.}
\label{tb:vit-hyperparam}
\end{table*}

\subsection{Supplements for method and analysis}

\subsubsection{Adam-{\ours} Algorithm}
\label{app:sage-adam}

\begin{algorithm}[htb!]
	\caption{Adam-{\ours} ($\odot$ denotes Hadamard product and $\oslash$ denotes Hadamard division)}
	\label{alg:main}
	\begin{algorithmic}[1]
		\INPUT Model parameters $\boldsymbol{\Theta} \in \RR^{J}$; Data $\cD$; Learning rate schedule $\eta(\cdot)$; Total training iteration $T$; Moving average coefficient $\beta_{0}, \beta_{1}, \beta_{2}$. \\
		Initialize $\hat{I}^{(0)}, m^{(0)}, v^{(0)}  = \boldsymbol{0} \in \RR^{J}$. 
		\For{$t = 1, ..., T$}
    		\State Sample a minibatch $b^{(t)}$ from $\cD$.
    		
    		\State Compute gradient $g^{(t)} = \nabla_{\boldsymbol{\Theta^{(t)}}} L(b^{(t)}, \boldsymbol{\Theta}^{(t)})$.
    		\State Compute sensitivity $I^{(t)} = |\boldsymbol{\Theta}^{(t)} \odot g^{(t)}|$.
    		\State $m^{(t)} = \beta_1 m^{(t-1)}  + (1 - \beta_1) g^{(t)}$
    		\State $v^{(t)} = \beta_2 v^{(t-1)}  + (1 - \beta_2) (g^{(t)})^{2}$
    		\State $\hat{I}^{(t)} = \beta_0 \hat{I}^{(t-1)} + (1-\beta_0) I^{(t)}$.
    		\State $\hat{m}^{(t)} = m^{(t)} / (1 - \beta_1)$
    		\State $\hat{v}^{(t)} = v^{(t)} / (1 - \beta_2)$
    		\State $\hat{I}^{(t)} = \hat{I}^{(t)} / (1 - \beta_0)$
		    \State $U^{(t)} = |I^{(t)} - \hat{I}^{(t)}|$.
    	    \State Update $\boldsymbol{\Theta}^{(t+1)} = \boldsymbol{\Theta}^{(t)} - \eta^{(t)} ((U^{(t)} + \epsilon) \odot \hat{m}^{(t)}) \oslash ((\hat{I}^{(t)} + \epsilon)\odot(\sqrt{\hat{v}^{(t)}} + \epsilon)) \odot g^{(t)}$.
		\EndFor
	\end{algorithmic}
\end{algorithm}

\subsubsection{Implementation Details for Section~\ref{sec:suff-train}}
\label{app:suff-train}

Figure~\ref{fig:ipt_dist} experiments: Due to the extremely large model size, we only sample $110$K parameters per layer (in total $12\times 110$K parameters) to calculate the distribution. We select the hyper-parameters that yield the best generalization performance on the BERT-base model, and we evaluate the sensitivity of each parameter using the entire training set.

Figure~\ref{fig:ipt_stability} experiments: Following previous experiment's practice, we randomly sample $110$K parameters per layer (in total $12\times 110$K parameters), and for visualization purposes, we plot $60$ randomly selected iterations. We adopt the learning rate corresponding to the best training performance for both SAGE and the baselines.

\subsubsection{Implementation Details for Section~\ref{sec:ana_generalization}}
\label{app:ana_generalization}
Plotting the parameter sensitivity distribution throughout training can be computational expensive. The distribution varies significantly throughout training and often fails to provide a meaningful visualisation. As a result, we compute the structured sensitivity score instead of the parameter sensitivity score. Specifically, we compute a single sensitivity score for each Transformer weight block $\Theta$ at iteration $t$ using the structured counterpart of the parameter sensitivity metric widely adopted in the existing structured pruning literature \citep{michel2019sixteen,liang2021super}. Following common structured pruning practice, we split Transformer models into $12$ feed-forward weight modules and $12$ multi-head attention weight modules, and plot the average and variance of the sensitivity of these modules' sensitivity scores throughout the training.

We present the results obtained with the hyper-parameters that yield the best generalization performance on the BERT-base model for both Adamax (Baseline) and Adamax-SAGE (SAGE).

\subsubsection{Ablation Study}
\label{app:ana_ablation}
To further interpret the role of the parameter sensitivity $I$ and the local temporal variation $U$, we conduct an ablation study on these two factors. Specifically, we check five variants of Eq.~(\ref{eq:alg}): 

\begin{gather*}
    \begin{split}
        \text{Variant 1.} \quad \eta^{(t)}_j  {}&= \eta^{(t)}  (\hat{I}_j^{(t)} + \epsilon)  (U_j^{(t)} + \epsilon) \\
        \text{Variant 2.} \quad \eta^{(t)}_j  &= \eta^{(t)}  (\hat{I}_j^{(t)} + \epsilon) /  (U_j^{(t)} + \epsilon)\\
        \text{Variant 3.} \quad \eta^{(t)}_j  &= \eta^{(t)}  (\hat{I}_j^{(t)} + \epsilon)\\
        \text{Variant 4.} \quad \eta^{(t)}_j  &= \eta^{(t)} / (\hat{I}_j^{(t)} + \epsilon) \\
        \text{Variant 5.} \quad \eta^{(t)}_j  &= \eta^{(t)}  (U_j^{(t)} + \epsilon)
    \end{split}
\end{gather*}

For Variants 1,2 and 3, we aim to check the performance of giving a high/low-sensitive parameter a high/low, instead of low/high learning rate. Specifically, we place $(\hat{I}_j^{(t)} + \epsilon)$ in the numerator, so that the learning rates increase for the high sensitive parameters and decrease for low sensitive parameters. 

For Variants 4 and 5, we aim to check the performance of eliminating the influence of one of these factors. Specifically, we fix the local temporal variation term to $1$ in Variant 4 and fix the sensitivity term to 1 in Variant 5. 

\subsubsection{Hyper-parameter Study}
\label{app:param_study}
\begin{figure}[htb!]
    \centering
    \vspace{-0.05in}
    \includegraphics[width=1.0\linewidth]{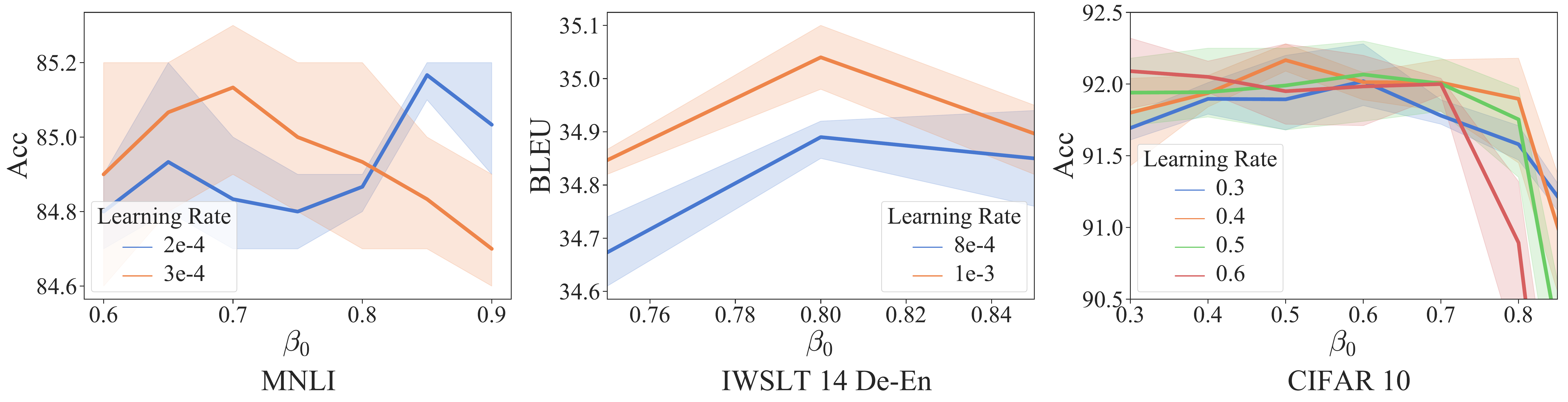}
	\caption{Parameter study on learning rate and $\beta_{0}$.}
	\label{fig:param_study}
	\vspace{-0.05in}
\end{figure}

We investigate the influence of hyper-parameters learning rate and $\beta_{0}$ on the performance of {\ours} (Figure~\ref{fig:param_study}). As can be seen, {\ours} requires a larger learning rate than the baselines to offset the small scale of the modulation term (the optimal baseline learning rate lies in $5 \times 10^{-5}\sim1\times 10^{-4}$ for MNLI, $5\times 10^{-4}\sim 7\times 10^{-4}$ for IWSLT 14 De-En and $0.1\sim 0.2$ for CIFAR10). Furthermore, switching to a larger learning rate requires a lower $\beta_{0}$ to maintain the same level of performance. 
% In all tasks, a change in $0.1$ in $\beta_{0}$ may result in a variance around $0.2$ points in accuracy, suggesting choosing a proper local window to compute the temporal variation is critical to success.

\newcolumntype{C}{@{\hskip4pt}c@{\hskip4pt}}
\begin{table*}[htb!]
\vspace{-0.05in}
\centering
\small
\begin{tabular}{l|l|CC}
\toprule
\textbf{Variant Name} & \textbf{Learning Rate Modulating Term} & \textbf{RTE} & \textbf{SST-2} \\ \midrule 
Adam       & $1$  & 63.5  &  92.9 \\ 
Adam-SAGE  & $(U_j^{(t)} + \epsilon) / (\hat{I}_j^{(t)} + \epsilon)$ & 73.3 & 93.5 \\ \midrule
Variant 1. & $(\hat{I}_j^{(t)} + \epsilon)(U_j^{(t)} + \epsilon)$ & 63.5 & 91.2 \\ 
Variant 2. & $(\hat{I}_j^{(t)} + \epsilon) / (U_j^{(t)} + \epsilon)$ & \text{Unconverged} & \text{Unconverged} \\
Variant 3. & $\hat{I}_j^{(t)} + \epsilon$  & 63.8 & 91.1 \\ 
Variant 4. & $1 /  (\hat{I}_j^{(t)} + \epsilon)$ & \text{Unconverged} & \text{Unconverged} \\ 
Variant 5. & $U_j^{(t)} + \epsilon$ & 63.8 & 91.1 \\ \bottomrule
\end{tabular} 
%\vspace{-0.05in}
\caption{Ablation study on parameter sensitivity and local temporal variations.}
\label{tb:}
\vspace{-0.05in}
\end{table*}

All five variants show no clear gain upon the baseline on both RTE and SST-2 datasets after careful hyper-parameter tuning. Specifically, we observe that the Variants 1 and 3 converge very fast at the early stage of training, and then quickly start overfitting. In Variants 2 and 4, the training collapses due to gradient explosion or vanishing.

\begin{table*}[htb]
	\begin{center}
		\begin{tabular}{l|l|c|c|c|c|c}
			\toprule 
			\bf Corpus &Task& \#Train & \#Dev & \#Test   & \#Label &Metrics\\ \midrule
			\multicolumn{6}{@{\hskip1pt}r@{\hskip1pt}}{Single-Sentence Classification (GLUE)} \\ \hline
			CoLA & Acceptability&8.5k & 1k & 1k & 2 & Matthews corr\\ \hline
			SST & Sentiment&67k & 872 & 1.8k & 2 & Accuracy\\ \midrule
			\multicolumn{6}{@{\hskip1pt}r@{\hskip1pt}}{Pairwise Text Classification (GLUE)} \\ \hline
			MNLI & NLI& 393k& 20k & 20k& 3 & Accuracy\\ \hline
            RTE & NLI &2.5k & 276 & 3k & 2 & Accuracy \\ \hline
            % WNLI & NLI &634& 71& 146& 2 & Accuracy \\ \hline
			QQP & Paraphrase&364k & 40k & 391k& 2 & Accuracy/F1\\ \hline
            MRPC & Paraphrase &3.7k & 408 & 1.7k& 2&Accuracy/F1\\ \hline
			QNLI & QA/NLI& 108k &5.7k&5.7k&2& Accuracy\\ \midrule
			\multicolumn{6}{@{\hskip1pt}r@{\hskip1pt}}{Text Similarity (GLUE)} \\ \hline
			STS-B & Similarity &7k &1.5k& 1.4k &1 & Pearson/Spearman corr\\ \bottomrule
% 			\multicolumn{6}{@{\hskip1pt}r@{\hskip1pt}}{Pairwise Text Classification} \\ \hline
% 			SNLI & NLI& 549k &9.8k&9.8k&3& Accuracy\\ \hline
% 			SciTail & NLI& 23.5k &1.3k&2.1k&2& Accuracy\\ \bottomrule
% 			ANLI & NLI& 163k &3.2k&3.2k&3& Accuracy\\ \hline
		\end{tabular}
	\end{center}
	\caption{Summary of the GLUE benchmark.}
	\label{tab:glue}
\end{table*}

\end{document}